\documentclass{article}

\PassOptionsToPackage{numbers, compress}{natbib}
\PassOptionsToPackage{dvipsnames, table}{xcolor}

\usepackage[preprint]{main}
\usepackage{tikz}
\usepackage{subcaption}
\usepackage{caption}
\usepackage{siunitx}
\usepackage{adjustbox}
\usepackage{tabularx}

\usepackage[utf8]{inputenc} 
\usepackage[T1]{fontenc}    
\usepackage{hyperref}       
\usepackage{url}            
\usepackage{booktabs}       
\usepackage{amsfonts}       
\usepackage{nicefrac}       
\usepackage{microtype}      

\usepackage{microtype}
\usepackage{graphicx}
\usepackage{booktabs} 
\usepackage{titlesec}

\usepackage{hyperref}
\usepackage{enumitem}
\usepackage{algorithm}
\usepackage{algpseudocode}
\usepackage{pifont}     

\usepackage[most]{tcolorbox}   

\usepackage{pifont}      
\usepackage{array,booktabs,graphicx}

\newcommand{\best}[1]{\cellcolor{gray!20}\textbf{#1}}

\newcommand{\cmark}{\textcolor{SkyBlue}{\ding{51}}}  
\newcommand{\xmark}{{\ding{55}}}

\setlist[itemize]{noitemsep, topsep=0pt}

\usepackage{todonotes}
\usepackage{url}
\usepackage{multirow}
\usepackage{arydshln} 
\usepackage{amssymb}
\usepackage{tikz}
\usepackage{wrapfig}
\usepackage{array}
\usepackage{float}
\usepackage{sidecap}

\usepackage{amsmath, bm}
\usepackage{amssymb}
\usepackage{mathtools}

\usepackage{listings}

\definecolor{codegreen}{rgb}{0,0.6,0}
\definecolor{codegray}{rgb}{0.5,0.5,0.5}
\definecolor{codepurple}{rgb}{0.58,0,0.82}
\definecolor{backcolour}{rgb}{0.95,0.95,0.92}

\lstdefinestyle{mystyle}{
    backgroundcolor=\color{backcolour},   
    commentstyle=\color{codegreen},
    keywordstyle=\color{magenta},
    numberstyle=\tiny\color{codegray},
    stringstyle=\color{codepurple},
    basicstyle=\ttfamily\footnotesize,
    breakatwhitespace=false,         
    breaklines=true,                 
    captionpos=b,                    
    keepspaces=true,                 
    numbers=left,                    
    numbersep=5pt,                  
    showspaces=false,                
    showstringspaces=false,
    showtabs=false,                  
    tabsize=2
}
\definecolor{templatebg}{HTML}{F7F9FC}     
\definecolor{titlegray}{RGB}{40, 40, 40}   
\definecolor{textblue}{RGB}{30, 90, 140}   
\definecolor{textgray}{RGB}{90, 90, 90} 

\title{First Finish Search: Efficient Test-Time Scaling in Large Language Models}

\author{
  Aradhye Agarwal \\
  Indian Institute of Technology Delhi \\
  \texttt{Aradhye.Agarwal.cs520@cse.iitd.ac.in} \\
  \And
  Ayan Sengupta \\
  Indian Institute of Technology Delhi \\
  \texttt{ayan.sengupta@ee.iitd.ac.in} \\
  \And
  Tanmoy Chakraborty \\
  Indian Institute of Technology Delhi \\
  \texttt{tanchak@iitd.ac.in} \\
}

\begin{document}

\maketitle

\begin{abstract}

Test‐time scaling (TTS), which involves dynamic allocation of compute during inference, offers a promising way to improve reasoning in large language models. While existing TTS methods work well, they often rely on long decoding paths or require a large number of samples to be generated, increasing the token usage and inference latency. We observe the surprising fact that \textit{for reasoning tasks, shorter traces are much more likely to be correct than longer ones}. Motivated by this, we introduce \textbf{First Finish Search (FFS)}, a training‐free parallel decoding strategy that launches $n$ independent samples and returns as soon as any one completes. We evaluate FFS alongside simple decoding, beam search, majority voting, and budget forcing on four reasoning models (DeepSeek-R1, R1-Distill-Qwen-32B, QwQ-32B and Phi-4-Reasoning-Plus) and across four datasets (AIME24, AIME25-I, AIME25-II and GPQA Diamond). 
With DeepSeek-R1, FFS achieves 82.23\% accuracy on the AIME datasets, a 15\% improvement over DeepSeek-R1's standalone accuracy, nearly matching OpenAI's o4-mini performance. Our theoretical analysis explains why stopping at the shortest trace is likely to yield a correct answer and identifies the conditions under which early stopping may be suboptimal. 
The elegance and simplicity of FFS demonstrate that straightforward TTS strategies can perform remarkably well, revealing the untapped potential of simple approaches at inference time.

\end{abstract}

\section{Introduction}

Large language models (LLMs)~\citep{yang2025qwen3technicalreport,guo2025deepseek,openai2024reasoning} have achieved impressive performance on complex reasoning tasks~\citep{huang2024olympicarena}, predominantly due to larger model sizes, greater pretraining compute, and vast training corpora~\citep{li2025system,bandyopadhyay2025thinking}. Further improvements, however, are proving increasingly difficult to achieve as the benefits of increasing model size diminish and the cost of further pretraining rises~\citep{zhang2025and}. Recent research has, instead, started exploring the possibility of test-time intervention~\citep{welleck2024decoding} for enhancing LLM performance. This approach, known as \textbf{test-time scaling (TTS)}~\citep{snell2024scaling}, dynamically allocates compute during decoding in order to improve accuracy and consistency without any additional training. TTS is especially effective on complex and reasoning-heavy benchmarks, where greater compute often translates into more reliable and precise outputs. OpenAI's o1 \citep{openai2024reasoning} and DeepSeek-R1 \citep{guo2025deepseek} are notable examples of models which utilize this approach.

According to the taxonomy proposed by \citet{zhang2025and}, TTS strategies fall into three main categories: parallel scaling~\citep{snell2024scaling,brown2024large}, sequential scaling~\citep{li2025s,li2025learning}, and hybrid approaches~\citep{xie2024monte, wang2025mcts}. Parallel scaling methods, such as beam search (BS), diverse beam search (DVBS) \citep{vijayakumar2016diverse}, and majority voting (MV), involve generating multiple candidate outputs concurrently and selecting the best response using heuristics or scoring functions. While these approaches excel in throughput and leverage computational parallelism effectively, they suffer from high token usage and require sophisticated mechanisms for evaluating and comparing candidate responses. For instance, MV assumes that the generated outputs can be easily compared using string equality or semantic similarity, which is often not the case in open-ended tasks. Sequential scaling techniques, on the other hand, extend the reasoning path deliberately with methods like budget forcing (BF)~\citep{muennighoff2025s1} introducing artificial delay tokens such as \textit{``Wait''} to encourage deeper reasoning, while others like thought switching penalty (TIP)~\citep{wang2025thoughts} discouraging premature shifts in reasoning direction. While these methods are powerful in improving accuracy, especially in tasks where extended deliberation leads to better answers, they are computationally expensive and introduce latency, making them less suitable for real-time or API-constrained deployments. Hybrid strategies, another class of TTS methods, combine elements of both parallel and sequential scaling to adapt to task difficulty and budget constraints. While combining diverse strategies does allow test-time compute to be allocated more judiciously, hybrid methods, which mix sequential and parallel strategies, still suffer from the same weaknesses to a greater or lesser extent.

\begin{wrapfigure}{r}{0.5\textwidth}
    \centering
    \includegraphics[width=\linewidth]{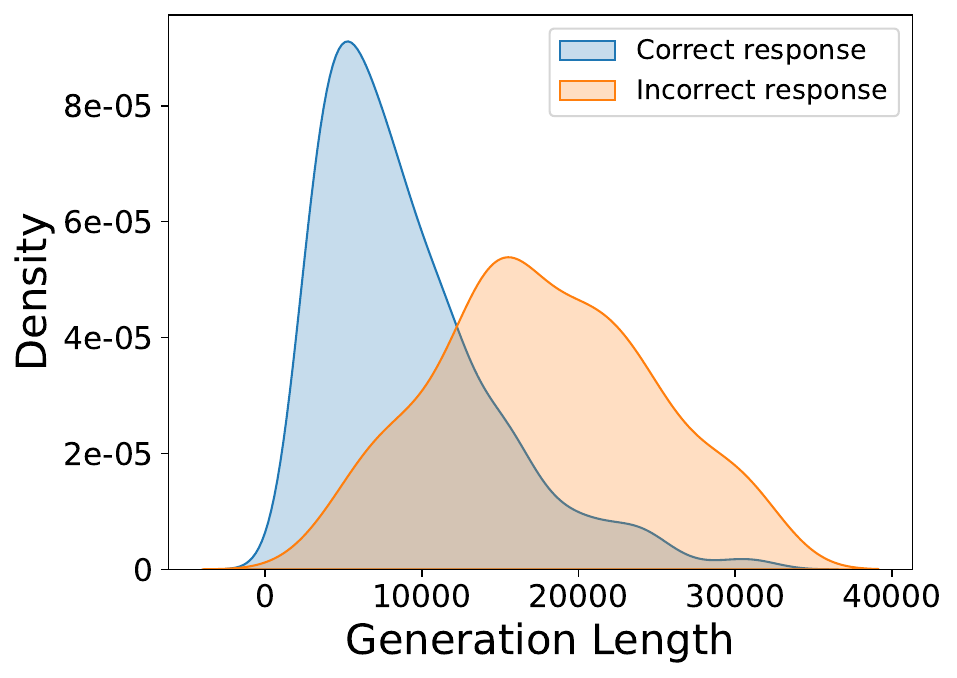}
    \caption{Distribution of trace lengths for correct and incorrect responses generated by QwQ~\citep{qwen2.5} and R1-Distill-Qwen models~\citep{guo2025deepseek}. The distributions show that shorter generation length and correctness are correlated, providing empirical justification for FFS. A Welch statistic of 16.56 with a p-value of $<$ 0.001 indicates the statistical significance of this behavior.}
    \label{fig:hypo1}
\end{wrapfigure}

To address these challenges, we introduce \textbf{F}irst \textbf{F}inish \textbf{S}earch \textbf{(FFS)}, a training-free test-time scaling method that launches $n$ samples in parallel and selects the output trace that completes first as the final answer. FFS is motivated by a surprising empirical finding (Figure \ref{fig:hypo1}) -- \textit{for reasoning tasks, shorter traces are significantly more likely to be correct}.\footnote{Note that Figure~\ref{fig:hypo1} actually implies the \textit{reverse}, i.e., correct traces are more likely to be short. We analyze in Appendix~\ref{reversehypothesis} as to why this also means that shorter traces are more likely to be correct.} FFS, unlike MV, needs no equality check between generated outputs, and unlike BF or TIP, it requires no special tokens or branching during decoding. It also uses far fewer tokens, and requires lesser sequential computation. We validate FFS with experiments on four benchmarks (GPQA Diamond, AIME24 and AIME25-I and AIME25-II) using four models: DeepSeek-R1, its distilled variant R1-Distill-Qwen-32B, QwQ-32B and Phi-4-Reasoning-Plus. Across all settings, FFS matches or exceeds the accuracy of strong baselines such as MV and BF while reducing token usage by up to 45\%. On AIME24 and AIME25 (AIME25-I and AIME25-II), with DeepSeek-R1, FFS attains an accuracy of 82.23\% on average, nearly reaching OpenAI's o4-mini (83.7\%) accuracy.\footnote{o4-mini accuracy is obtained from \texttt{https://www.vals.ai/benchmarks/aime-2025-05-09}} Our experiments show that FFS gains more from increased model capacity than any competing method, enabling it to outperform every baseline on DeepSeek-R1. Our theoretical results show that FFS's expected sequential cost decreases with increasing sample count, demonstrating its compute efficiency for inference-time scaling. These findings establish FFS as a scalable, efficient and effective alternative for boosting LLM reasoning at test-time.\footnote{The source code is available at \texttt{https://github.com/Aradhye2002/reasoning\_exps}}

\section{Related Work}

\begin{table}[t]
  \caption{Comparison of test-time scaling strategies across five criteria. ``Training-free'' indicates no additional model training is required; ``API-friendly'' means the method can be implemented using standard inference APIs without logit edits or repeated calls; ``scalable'' denotes the ability to improve performance by allocating more compute; ``T-parallelizable'' means total compute cost decreases when decoding is parallelized across more workers; and ``S-parallelizable'' means sequential latency decreases as the number of parallel samples increases. \textbf{DVBS:} diverse beam search, \textbf{BS:} beam search, \textbf{TIP:} thought switching penalty, \textbf{BF:} budget forcing, \textbf{SD:} simple decoding, \textbf{MV:} majority voting, \textbf{LFS:} last finish search, \textbf{FFS:} first finish search.}
  \vspace{1em}
  \label{tab:method-comparison}
  \centering
  \setlength{\tabcolsep}{8pt}      
  \renewcommand{\arraystretch}{1.1}
  \begin{tabular}{lccccccccc}
    \toprule
    & DVBS & BS & TIP & BF & SD & MV & LFS & L1 & FFS \\
    \midrule
    Training-free      & \cmark & \cmark & \cmark & \cmark & \cmark & \cmark & \cmark & \xmark & \cmark \\
    API-friendly       & \xmark & \xmark & \xmark & \xmark & \cmark & \cmark & \cmark & \cmark & \cmark \\
    Scalable           & \cmark & \cmark & \xmark & \cmark & \xmark & \cmark & \cmark & \xmark & \cmark \\
    T-parallelizable   & \cmark & \cmark & \xmark & \xmark & \xmark & \cmark & \cmark & \xmark & \cmark \\
    S-parallelizable   & \xmark & \xmark & \xmark & \xmark & \xmark & \xmark & \xmark & \xmark & \cmark \\
    \bottomrule
  \end{tabular}
\end{table}

TTS techniques aim to improve LLM reasoning by dynamically adjusting the compute budget~\citep{snell2024scaling} during inference, without needing expensive model fine-tuning. Existing TTS methods can be categorized into two broad categories:

\textbf{1. Training‐based strategies.} In these methods, the base model is fine‐tuned in order to \emph{follow} a particular compute‐allocation policy at inference time.  For instance, Inference‐Aware Fine‐Tuning (IA-FT)~\citep{chow2024inference} optimizes a supervised or RL objective directly targeting best-of-N accuracy: during pretraining, the model learns to generate diverse candidates so that pass@N is optimized.  Similarly, L1~\citep{aggarwal2025l1} trains the model to obey an explicit ``think for $L$ tokens'' instruction in the prompt, granting the user direct control over the length of the chain-of-thought.  While powerful, these methods require access to model weights, substantial compute for fine‐tuning, and are not applicable to closed‐source or API‐only LLMs.

\textbf{2. Training‐free strategies.}  
Instead of fine-tuning the existing model weights, these methods work by adapting additional compute-allocation at decoding time. Based on the decoding strategies, these methods can be further classified into: \textbf{parallel scaling}, where decoding threads are launched in parallel from the same prompt, and a selection rule picks one final output.  Examples include BS, which maintains a fixed‐width beam of top-scoring partial sequences; DVBS, which imposes diversity penalties to encourage variation before ranking; and MV/self-consistency, which samples $N$ full answers and returns the one with highest agreement. Other notable parallel scaling strategies~\citep{levi2024simple,peng2025agentic,zhang2025and} utilize the best response based on an external reward models. In contrary to parallel scaling methods, in another class of training-free TTS approach \textbf{sequential scaling}, a single decoding thread deliberately extends model's reasoning chain to consume more tokens before yielding an answer~\citep{hou2501advancing,lee2025evolving}. Budget forcing interposes special ``wait'' tokens to delay termination artificially, while the TIP applies dynamic logit penalties to discourage early topic shifts. Recently proposed tree-based TTS methods~\citep{zhang2023planning,wu2024empirical,gandhi2024stream,yang2025step} draw balance between sequential and parallel scaling strategies. \textbf{Hybrid scaling} methods like Monte Carlo Tree Search (MCTS)~\citep{liu2023don,pan2025coat} and self-backtracking~\citep{yang2025step} aim to recognize and correct suboptimal reasoning paths through simultaneous parallel and sequential decoding.

\noindent This taxonomy highlights a clear operational divide: training‐based approaches encode compute decisions into the model parameters, while training‐free methods manipulate the decoding process itself.  Within the latter, one must choose between spawning many concurrent trajectories or elongating a single trajectory, each with distinct implications. We evaluate these methods along five dimensions:

\begin{itemize}[leftmargin=1em]
\item \textbf{Training‐free}: Does the method require any fine‐tuning or reinforcement learning?
\item \textbf{API‐friendly}: Can the method be implemented using only standard model‐serving API?
\item \textbf{Scalable}: Can performance continue to improve as we allocate more compute (more samples or longer generations)?
\item \textbf{T‐parallelizable}: Does total GPU/CPU cost drop when decoding is parallelized across more workers?
\item \textbf{S‐parallelizable}: Does the sequential latency (time to first answer) decrease when we parallelize across more samples?
\end{itemize}

\begin{figure}[t]
  \centering
  \includegraphics[width=\linewidth]{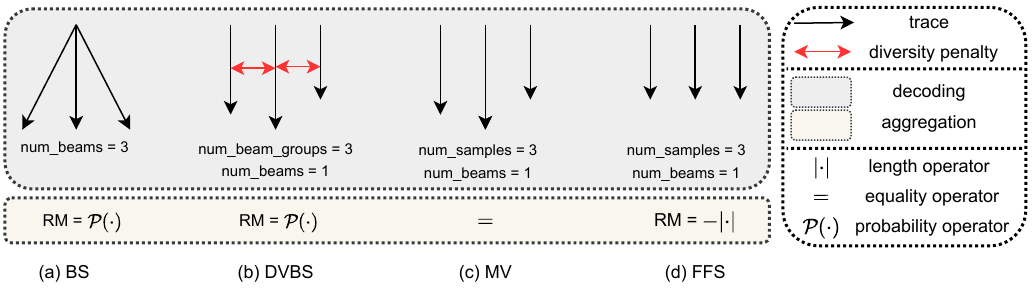}
  \vspace{-0.3em}
  \caption{Sampling-based TTS strategies.
           (a) BS:
           expands \(k\) partial hypotheses synchronously, ranking both
           intermediate and final beams with the model probability
           \(\mathcal{P}(\cdot)\).  Beams share prefixes, so their
           lengths tend to remain aligned.  (b)
           DVBS: launches \(g\) independent
           single-beam groups; a diversity term keeps the groups apart,
           so they may hit EOS at different steps.  After all
           groups finish, a verifier (again \(\mathcal{P}(\cdot)\) here)
           selects the best trace.  (c)~MV/self-consistency: draws \(n\) complete,
           independent samples and picks the modal answer using a string
           equality test ``\(=\)''.  (d) FFS: starts the same \(n\) stochastic samples but
           terminates the batch as soon as the first trace reaches
           EOS; all displayed arrows are therefore cut to that
           minimal length.  The selection rule is equivalent to
           maximising the negative length reward
           \(R(T) = -|T|\).}
  \label{fig:parallel_tts}
\end{figure}

Table~\ref{tab:method-comparison} summarizes how each method performs across key criteria. BS, DVBS, and MV support parallel scaling by generating multiple outputs and selecting the best (Figure~\ref{fig:parallel_tts}); they are training-free and T-parallelizable, but BS and DVBS are not API-friendly due to requiring custom diversity penalties or verifier models, while MV is API-compatible but not S-parallelizable, as it must wait for all outputs before voting. BF and TIP scale sequentially by encouraging longer reasoning, are training-free, and potentially API-friendly if token manipulation is allowed, but repeated API calls make BF practically infeasible; neither is parallelizable and both are limited by model context length. Training-based methods like supervised fine-tuning and RL approaches (e.g., IA-FT, L1) improve accuracy but require model access and heavy compute, are API-incompatible, and are not parallelizable beyond the model's sequence limits. FFS combines the best of both: it is training-free, API-friendly, uses standard sampling with a stop-when-finished rule, scales arbitrarily with more samples, is T-parallelizable (lower cost with parallel runs), and uniquely S-parallelizable (latency drops as shorter correct traces finish early), making it a lightweight and effective alternative.

\begin{algorithm}[!t]
\caption{Synchronous FFS (Sync-FFS)}
\label{alg:sync-ffs}
\begin{algorithmic}[1]
\Require Reasoning model $\mathcal{M}$, prompt $x$, number of traces $n$, max length $L$
\State Initialize partial traces $T_1,\dots,T_n \leftarrow \text{[BOS]} \circ x$
\For{$\ell = 1$ \textbf{to} $L$}
    \State $\mathbf{Y} \leftarrow \mathcal{M}(T_1,\dots,T_n)$ \Comment{one batched forward pass}
    \For{$i = 1$ \textbf{to} $n$}      \Comment{sample next token for each trace}
        \State sample $y_i \sim \text{softmax}(\mathbf{Y}_i)$
        \State $T_i \leftarrow T_i \circ y_i$
        \If{$y_i = \textsc{EOS}$}
            \State $T^\star \leftarrow T_i$  \Comment{first finished trace}
            \State \Return $T^\star$
        \EndIf
    \EndFor
\EndFor
\State \Return $T_1$  \Comment{fallback if no trace hit \textsc{EOS}}
\end{algorithmic}
\end{algorithm}

\begin{algorithm}[!th]
\caption{Asynchronous FFS (Async-FFS)}
\label{alg:ffs}
\begin{algorithmic}[1]
\Require Reasoning model $\mathcal{M}$, prompt $x$, number of traces $n$, max decode length $L$
\State \textbf{Launch} $n$ asynchronous decoding jobs $\{J_1,\dots,J_n\}$, each fed with $x$
\While{true}
    \ForAll{running jobs $J_i$ \textbf{in parallel}}
        \If{$J_i$ produces \textsc{EOS} \textbf{or} $|T_i| = L$}
            \State $T^\star \leftarrow T_i$ \Comment{store the first finished trace}
            \State \textbf{Interrupt} all jobs $J_{k\neq i}$ and free their resources
            \State \Return $T^\star$
        \EndIf
    \EndFor
\EndWhile
\end{algorithmic}
\end{algorithm}
\vspace{5pt}

\section{FFS: Our Proposed Method}

FFS runs $n$ independent decoding operations in parallel and stops as soon as any one trace emits the end-of-sequence token. The winning trace is returned and all others are discarded. In order to ensure diversity among the generations, FFS uses stochastic decoding with beam size 1.

FFS admits two implementations -- Sync-FFS and Async-FFS, which constitute the synchronous and asynchronous variants of the high-level algorithm, respectively. Sync-FFS (Algorithm \ref{alg:sync-ffs}) loads a single copy of the model $\mathcal M$ and processes a batch of $n$ partial sequences in lock-step. At every decoding step we sample one token for each partial sequence; if any of those tokens is an EOS symbol, we immediately terminate decoding and return that completed trace. By sharing the model across all samples, Sync-FFS minimizes total compute and memory, making it well suited for centralized servers or GPUs. Async-FFS (Algorithm \ref{alg:ffs}) launches $n$ independent decoding jobs, on separate processes or machines, all starting from the same prompt. When any job produces EOS (or reaches the token limit), it interrupts the remaining $n-1$ jobs, frees their resources, and returns the completed trace. This variant naturally fits distributed or multi-worker environments.

While our motivating hypothesis states that shorter reasoning segments are likely to be more accurate, we simplify implementation by measuring the entire trace length (reasoning plus final answer). Since the solution portion is typically small, ranking by full length almost always aligns with ranking by reasoning length. Also, while larger beams can reduce repetition, we deliberately use beam size 1 in order to maximize the number of independent samples, and consequently the chance that a short, correct trace finishes first. In practice, repetitive or degenerate beams rarely finish earliest, so they are automatically filtered out by FFS's ``first-to-finish'' rule.

To back our claim that shorter traces are likelier to be correct, we derive the following expression for the probability that a randomly drawn trace of length $x$ is correct.

\textbf{Result 1.} Suppose correct and incorrect samples are drawn from distinct normal distributions with means $\mu_1$, $\mu_2$ and standard deviations $\sigma_1$, $\sigma_2$ respectively. Let $\alpha$ be proportion of samples that are correct. Then the probability that a randomly sampled trace $T$ of a given length $x$ is correct is:
\begin{equation}
\label{eq:1}
\textrm{Pr}\Big[T \textrm{ is correct} \mid |T| = x\Big] = \left(1+\frac{1-\alpha}{\alpha}\frac{\sigma_1}{\sigma_2}e^{-\dfrac{1}{2}\left[\left(\dfrac{x-\mu_2}{\sigma_2}\right)^{\!\!2} - \left(\dfrac{x-\mu_1}{\sigma_1}\right)^{\!\!2}\right]}\right)^{\!\!-1}
\end{equation}

\begin{figure*}
    \centering
    \subfloat[Distribution of generation length for different AIME24 examples]{\includegraphics[width=0.6\linewidth]{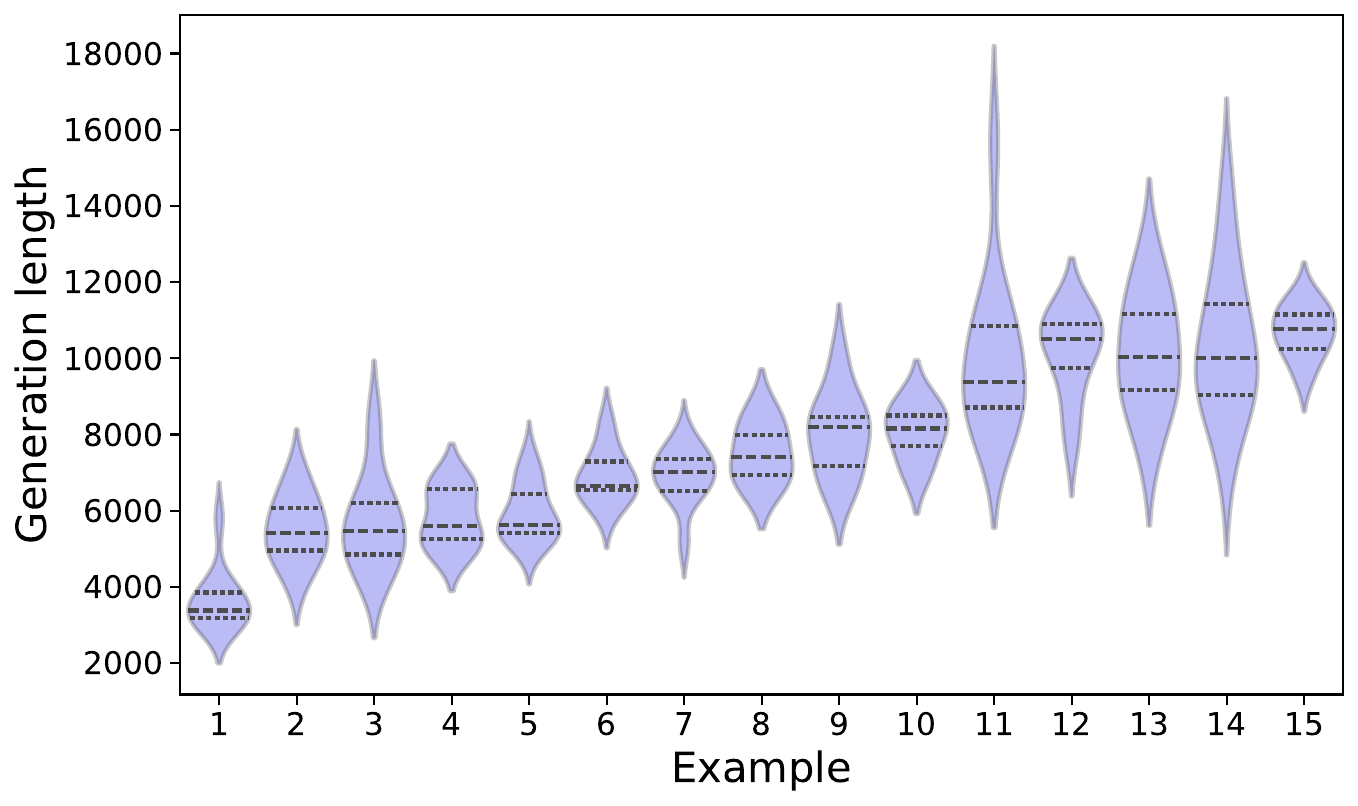}}
    \subfloat[P-value of normality test]{\includegraphics[width=0.4\linewidth]{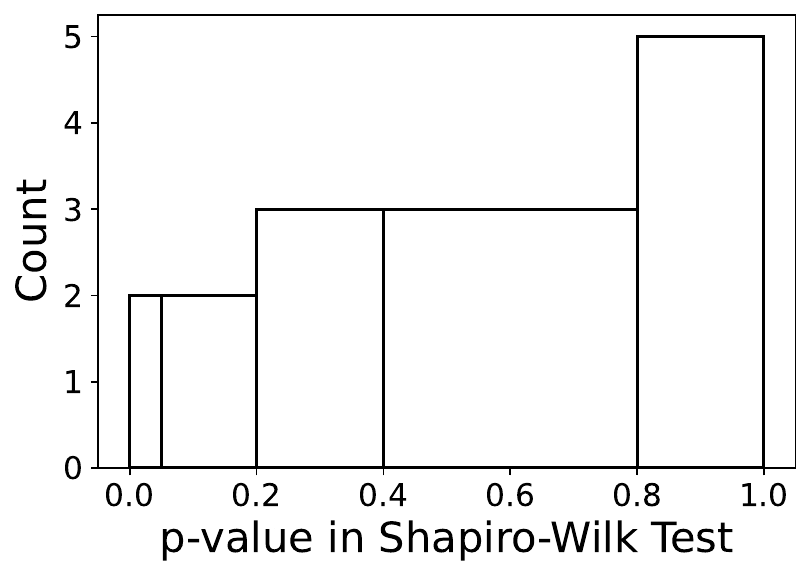}}
    \caption{Lengths of generated reasoning traces on AIME24. (a) Per-example violin plots of trace length over 8 samples, for each question. (b) Histogram of Shapiro-Wilk p-values for normality across examples, indicating that most trace-length distributions do not reject the gaussian assumption.}
    \label{fig:length_dist}
\end{figure*}

Since FFS always favours the shortest trace, we first examine the limiting case $x = 0$ in Equation \eqref{eq:1}.  With comparable variances $(\sigma_{1}\!\approx\!\sigma_{2})$, the simplified form shows that the probability of correctness is high only when the mean length of incorrect traces exceeds that of correct ones, i.e.,\ $\mu_{2} > \mu_{1}$.  Empirically, reasoning models satisfy this inequality, confirming the link between conciseness and accuracy. A parallel ``long-trace'' analysis for LFS $(x \to \infty)$ is not meaningful.  Beyond moderate lengths the empirical trace-length distributions become heavy-tailed (Figure \ref{fig:length_dist}) and depart from the normal assumption that underlies Equation \eqref{eq:1}.
This analysis reveals that correctness of a trace is closely linked to how short it is, making it crucial to devise ways in order to discover increasingly shorter traces, with FFS being one such effective way.

To investigate how FFS's cost reduces with increasing number of samples $n$, we introduce the following result from extreme value theory~\citep{galambos1977asymptotic}, which states that the difference between the mean and expected minimum of $n$ i.i.d.\ normally distributed RVs is $\mathcal{O}(\sqrt{\log{n}})$. 

\textbf{Result 2.} Let $Y_1, Y_2, \dots, Y_n$ be i.i.d.\ normally distributed random variables with mean $\mu$ and standard deviation $\sigma$, then as $n \rightarrow \infty$, we have\footnote{See Appendices~\ref{result1proof} and ~\ref{result2proof} for detailed proofs.}
\begin{equation}
\label{eq:2}
\mathbb{E}\bigl[\min\{Y_1, Y_2, \dots, Y_n\}\bigr] = \mu - \sigma \sqrt{2\log{n}}
\end{equation}
\begin{equation}
\label{eq:3}
\mathbb{E}\bigl[\max\{Y_1, Y_2, \dots, Y_n\}\bigr] = \mu + \sigma \sqrt{2\log{n}}
\end{equation}

In order to use Result 2 in our analysis of FFS, we define $Y_i$ as $|T_i|$, where $T_i$ is a trace sampled from the model $\mathcal{M}$, and $|\cdot|$ is the usual length operator. 
Since $T_i$'s are independently and identically sampled, $Y_i = |T_i|$ are i.i.d.\ random variables. $Y_i$'s are also normally distributed for reasoning models as demonstrated through Figure~\ref{fig:length_dist}. On application of Equation~\eqref{eq:2}, we obtain that the expectation of the minimum trace length (which is same as FFS's sequential cost) decreases in $\mathcal{O}(\sqrt{\log{n}})$, with increasing $n$. Whereas, the maximum trace length (same as MV's sequential cost) increases with $n$.

\section{Experimental Setup}

We evaluate FFS with four reasoning models: Phi-4-Reasoning-Plus~\citep{abdin2025phi}, DeepSeek-R1~\citep{guo2025deepseek}, QwQ-32B~\citep{qwq32b} and R1-Distill-Qwen~\citep{qwen2.5} and one non-reasoning model: DeepSeek-V3~\citep{liu2024deepseek}. On DeepSeek-R1 and Phi-4-Reasoning-Plus, we evaluate only MV, LFS, and FFS, omitting BF and BS due to API constraints. On QwQ-32B and R1-Distill-Qwen, we report results for all five methods: MV, LFS, FFS, BF, and BS. Last finish search (LFS) is an additional baseline added by us for comparison with FFS, where we take the last finished trace. We conduct evaluations on four datasets: GPQA Diamond \citep{rein2024gpqa}, AIME24~\citep{aime}, AIME25-I and AIME25-II. The AIME datasets (AIME24, AIME25-I, and AIME25-II) totaling 60 problems, are derived from the American Invitational Mathematics Examination, consisting of challenging high school-level competition math problems. Each question has a numeric answer between 000 and 999. For consistent evaluation, we strip visual figures and convert all problems into text, consistent with the approach of \citep{muennighoff2025s1}. The GPQA Diamond dataset comprises 198 graduate-level multiple-choice science questions spanning physics, chemistry, and biology. Each question is designed to be google-proof and tests high-level conceptual reasoning. The task templates and output parsing logic for these datasets are provided in Appendix~\ref{appx:dataset}.

We evaluate each method using three metrics: (1) \textbf{accuracy}, measured as the proportion of exact matches with ground-truth answers; (2) \textbf{total compute}, defined as the total number of generated tokens across all parallel traces for a given question; and (3) \textbf{sequential compute}, which is the minimum number of sequential tokens required to be produced to obtain the final answer. For GPQA, we consider a response correct only if the selected option exactly matches the correct choice while for the three AIME datasets, only the last integer matches are considered. 

All models are evaluated in a zero-shot setting with no additional fine-tuning.\footnote{We use \texttt{deepinfra.com} API for all our evaluations.} We use \texttt{top\_p} = 0.95 and \texttt{temperature} = 0.6 for all methods across all datasets. For all sampling based methods (FFS, LFS, and MV) we take the number of samples \texttt{n} = 4. Since AIME problems require deeper reasoning, we set a maximum generation length of 32K tokens; on GPQA Diamond, 16K tokens suffice. Appendix~\ref{hyperparams_section} provides the complete hyperparameter list.

\section{Results}
\label{results}

\begin{table*}[t]
\centering
\caption{Accuracy (\%) and compute cost ($\times 10^{3}$ tokens) for
R1-Distill-Qwen and QwQ-32B.  
Rows list metrics; columns list decoding methods.  For each method, token counts
are averages over the datasets.  Bold, gray cells mark the best value in each
row.}
\label{tab:tts-results-1-fancy}
\setlength{\tabcolsep}{5pt}
\renewcommand{\arraystretch}{1.12}

\resizebox{\linewidth}{!}{
\begin{minipage}[t]{0.5\linewidth}
\centering
\caption*{(a) R1-Distill-Qwen}
\resizebox{\linewidth}{!}{
\begin{tabular}{lcccccc}
\toprule
Metric & SD & BF & BS & MV & LFS & FFS \\
\midrule
Seq.\ tokens  & 11.4 & 25.7 & 11.2 & 15.7 & 15.7 & \best{7.8} \\
Total tokens  & \best{11.4} & 25.7 & 44.8 & 45.8 & 45.8 & 31.3 \\
\hdashline
GPQA      & 60.7 & 58.6 & 62.6 & 62.1 & 60.1 & \best{62.6} \\
AIME24    & 68.3 & 60.0 & 66.7 & 77.9 & 60.6 & \best{80.0} \\
AIME25-I  & 51.7 & 53.3 & 46.7 & 53.3 & 49.7 & \best{59.6} \\
AIME25-II & 46.7 & 57.1 & 57.1 & \best{60.0} & \best{60.0} & 46.7 \\
\bottomrule
\end{tabular}}
\end{minipage}\hspace{1em}
%
\begin{minipage}[t]{0.5\linewidth}
\centering
\caption*{(b) QwQ-32B}
\resizebox{\linewidth}{!}{
\begin{tabular}{lccccccc}
\toprule
Metric & SD & BF & BS & MV & LFS & FFS \\
\midrule
Seq.\ tokens  & 14.9 & 23.7 & 12.8 & 18.8 & 18.8 & \best{11.8} \\
Total tokens & \best{14.9} & 23.7 & 51.2 & 59.7 & 59.7 & 47.2 \\
\hdashline
GPQA      &  -- & 60.1 & 57.1 & 64.7 & 57.6 & \best{65.2} \\
AIME24    & 79.2 & \best{86.7} & 80.0 & 83.5 & 80.1 & 81.2 \\
AIME25-I  & 60.0 & 60.0 & 66.7 & \best{69.9} & 53.5 & 59.9 \\
AIME25-II & 69.2 & 71.4 & 78.6 & \best{81.9} & 70.4 & 78.0 \\
\bottomrule
\end{tabular}}
\end{minipage}}
\end{table*}

\begin{table*}[t]
\centering
\caption{Accuracy (\%) and compute cost ($\times 10^{3}$ tokens) for
DeepSeek-R1, Phi-4-Reasoning-Plus, and DeepSeek-V3.  For each method, token
counts represent averages over all datasets.  Bold, gray cells mark the best
value per row.}
\label{tab:tts-results-2-fancy}
\setlength{\tabcolsep}{3pt}
\renewcommand{\arraystretch}{1.08}

\begin{tabular}{@{}c@{}}
\begin{minipage}[t]{0.33\linewidth}
\centering\small
\caption*{(a) DeepSeek-R1}
\resizebox{!}{0.33\linewidth}{
\begin{tabular}{lcccc}
\toprule
Metric & SD & MV & LFS & FFS \\
\midrule
Seq.\ tokens & 10.6 & 13.8 & 13.8 & \best{7.8} \\
Total tokens & \best{10.6} & 42.2 & 42.2 & 31.1 \\
\hdashline
GPQA      & 72.0 & 73.2 & 72.2 & \best{74.2} \\
AIME24    & 75.0 & 83.3 & 70.0 & \best{86.7} \\
AIME25-I  & 51.7 & 60.0 & 53.3 & \best{66.7} \\
AIME25-II & 75.0 & 80.0 & 60.0 & \best{93.3} \\
\bottomrule
\end{tabular}}
\end{minipage}
\hfill
\begin{minipage}[t]{0.33\linewidth}
\centering\small
\caption*{(b) Phi-4-Reasoning-Plus}
\resizebox{!}{0.33\linewidth}{
\begin{tabular}{lcccc}
\toprule
Metric & SD & MV & LFS & FFS \\
\midrule
Seq.\ tokens & 14.7 & 18.7 & 18.7 & \best{11.2} \\
Total tokens & \best{14.7} & 58.8 & 58.8 & 44.8 \\
\hdashline
GPQA      & 62.9 & 67.7 & \best{70.2} & 67.2 \\
AIME24    & 71.7 & \best{80.0} & \best{80.0} & 76.7 \\
AIME25-I  & 66.7 & 73.3 & \best{80.0} & 66.7 \\
AIME25-II & 76.7 & \best{93.3} & 86.7 & 86.7 \\
\bottomrule
\end{tabular}}
\end{minipage}
\hfill
\begin{minipage}[t]{0.33\linewidth}
\centering\small
\caption*{(c) DeepSeek-V3}
\resizebox{!}{0.33\linewidth}{
\begin{tabular}{lcccc}
\toprule
Metric & SD & MV & LFS & FFS \\
\midrule
Seq.\ tokens & 2.7 & 4.4 & 4.4 & \best{1.4} \\
Total tokens & \best{2.7} & 10.8 & 10.8 & 5.5 \\
\hdashline
GPQA      & 53.3 & \best{54.0} & 52.5 & 50.0 \\
AIME24    & 30.0 & \best{40.0} & \best{40.0} & 23.3 \\
AIME25-I  & 33.3 & \best{46.7} & 33.3 & \best{46.7} \\
AIME25-II & 16.7 & \best{20.0} & \best{20.0} & \best{20.0} \\
\bottomrule
\end{tabular}}
\end{minipage}
\end{tabular}
\end{table*}

\paragraph{Token-efficient accuracy gains.}
Across all evaluated settings FFS either matches or improves on baseline accuracy while reducing token usage (Tables \ref{tab:tts-results-1-fancy}-\ref{tab:tts-results-2-fancy}). With DeepSeek-R1 it scores 86.7\% on AIME24 and 93.3\% on AIME25-II while averaging only 31.1K total tokens and 7.8K sequential tokens per query, about 26\% less compute than MV's 42.2 K-token budget. On the smaller R1-Distill-Qwen-32B model, FFS reaches 80.0\% on AIME24 and 62.6\% on GPQA using the same 7.8 K/31.3K token budget, whereas MV consumes 15.7K/45.8K tokens for slightly lower scores. BS can equal FFS on a few cases (e.g.\ 66.7\% on AIME25-I) but does so at roughly 1.4 times the compute cost.

\paragraph{Consistent improvements across models and datasets.}
FFS remains competitive even on larger models. On QwQ-32B it reaches 78.0\% on AIME25-II, just 3.9 points less than MV's top score while using 25\% fewer tokens (47.2K vs.\ 59.7K). For Phi-4-Reasoning-Plus, FFS increases GPQA accuracy to 67.2\%, beating MV by 4.5 points while reducing the total token budget by 24\% (44.8K vs.\ 58.8K). On math-heavy AIME25-II it still attains 86.7\%, matching MV's peak performance but at a lower cost. In the most compute-intensive setting of DeepSeek-R1 on AIME25-II FFS outperforms LFS by 33\% and beats MV by 13\% while remaining the cheapest of the multi-sample strategies.

\paragraph{Linear scalability and reduced latency.}
Since FFS halts decoding once the shortest trace finishes, its cost scales linearly with the number of parallel samples and never exceeds a single-trace budget. Concretely, FFS reduces the sequential budget down to 7.8K tokens for DeepSeek-R1 and 11.8K for QwQ-32B, compared with 13-19K for MV and LFS. These savings translate directly into lower end-to-end latency in throughput-bound or API-metered deployments.

\paragraph{Behaviour on a non-reasoning model.}
On DeepSeek-V3, a model without explicit chain-of-thought supervision, we observe a reverse trend: MV surpasses both LFS and FFS on every dataset, and FFS records the lowest scores (e.g.\ 50\% on GPQA and 20\% on AIME25-II). This suggests that the ``shortest correct trace'' bias exploited by FFS is specific to models that already internalise multi-step reasoning.

\paragraph{FFS benefits from model scaling.}
The advantage of FFS widens with capacity. Upgrading from Distill-Qwen-32B to DeepSeek-R1 boosts the method's AIME accuracy from 80.0\% to 86.7\% on AIME24 and from 62.6\% to 74.2\% on GPQA, while preserving its compute edge. FFS attains 93.3\% on AIME25-II, over 14 points higher than the strongest baseline, showing that it scales more favourably with model size than competing approaches.

\section{Analysis}

\begin{wrapfigure}{r}{0.5\linewidth}   
  \centering
  \vspace{-15pt}
  \includegraphics[width=\linewidth]{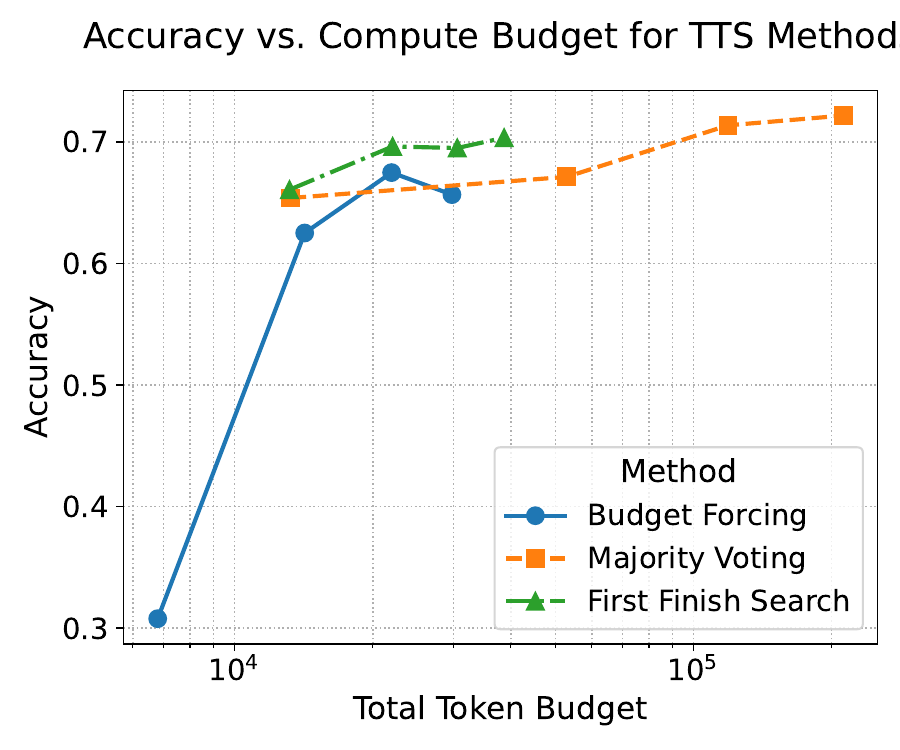}
  \caption{Accuracy versus total token budget for three test-time scaling methods averaged over R1-Distill-Qwen and QwQ-32B on AIME24 and AIME25-I. FFS attains higher accuracy at lower budgets; MV improves more slowly and consumes more tokens; BF plateaus near the 32K token limit.}
  \vspace{-12pt}
  \label{fig:tts_scaling}
\end{wrapfigure}

\paragraph{Scaling behaviour of TTS methods.}
Figure~\ref{fig:tts_scaling} plots accuracy versus total token budget, averaged over R1-Distill-Qwen and QwQ-32B on the AIME24 and AIME25-I tasks. BF attains roughly 30\% accuracy at the smallest budget and then plateaus once the generation reaches the 32K token limit. Its final point drops slightly because traces that hit this limit generally yield no answer and are marked incorrect. MV and FFS coincide at the single-sample setting, both achieving about 60\% accuracy, since FFS degenerates to MV when $n=1$. As the number of samples increases, MV’s accuracy improves slowly but at the cost of a steep rise in tokens, proportional to the mean trace length times the sample count. FFS, in contrast, halts decoding as soon as the first trace finishes, reaching comparable or higher accuracy with far fewer tokens; its curve therefore rises more sharply. Across the entire budget range, FFS matches or surpasses MV while operating at a lower cost.




\begin{figure}
    \centering
    \subfloat[R1-Distill-Qwen]{\includegraphics[width=0.33\linewidth]{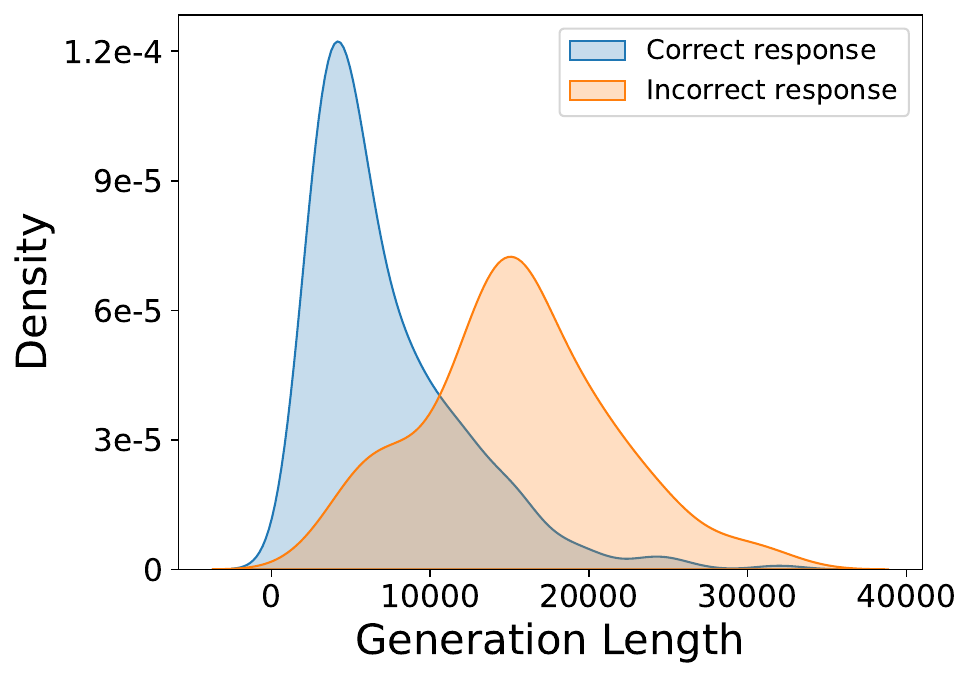}}
    \subfloat[QwQ-32B]{\includegraphics[width=0.33\linewidth]{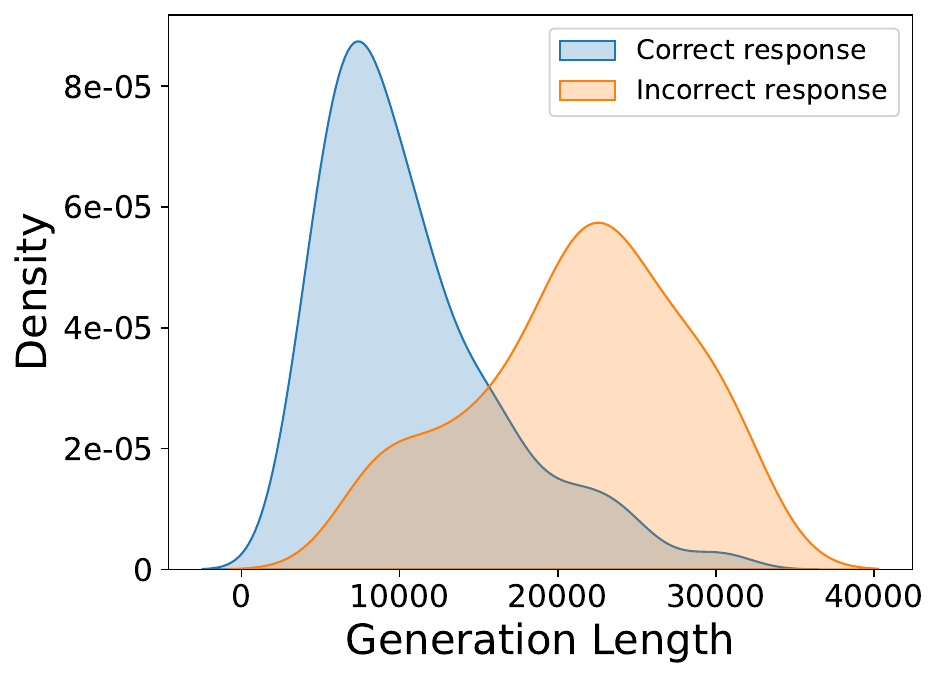}}
    \subfloat[DeepSeek-V3]{\includegraphics[width=0.33\linewidth]{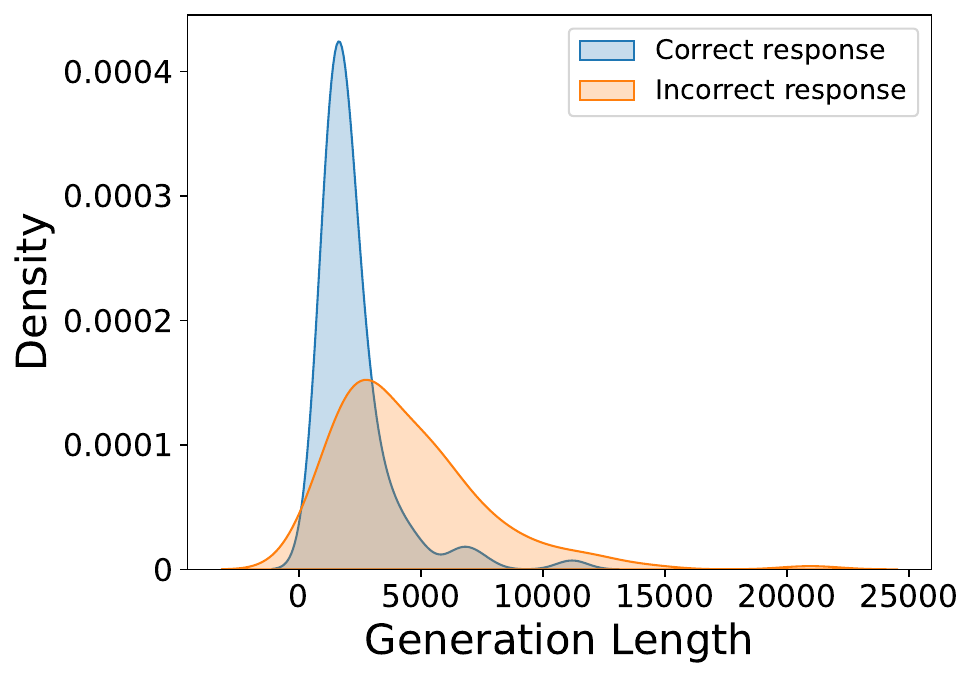}}
    \caption{Distribution of generated text lengths for different reasoning and non-reasoning models on AIME24 and AIME25 datasets. Welch statistic of 13.53, 13.3 and 6.9 with p-values $<$ 0.001 indicates the statistical significance of the fact that correct traces are more likely to be shorter.}
    \label{fig:hypothesis1_modelwise}
\end{figure}

\paragraph{A theoretical perspective on FFS’s scaling behavior with model size.}
As demonstrated in Section~\ref{results}, FFS shows greater relative improvements on the more capable DeepSeek-R1 model compared to other baselines. This aligns with our intuition: all traces begin in a correct state by default (no reasoning has occurred yet). Consider now an idealized oracle model that never transitions from a correct reasoning state to an incorrect one. In this case, any trace generated would either be entirely correct or remain incomplete. Therefore, among all correct traces, the shortest one would always be optimal. As model capacity increases and approaches oracle-like behavior, the likelihood of making an incorrect reasoning step diminishes. Consequently, by favoring shorter correct traces, FFS naturally benefits from the improved reliability of larger models, leading to its enhanced performance.

\begin{wraptable}{r}{0.6\linewidth}
  \centering
  \caption{Trace-length statistics and predicted FFS success rates, measured per 1000 tokens. Averages are computed over the AIME datasets.} 
  \label{tab:length-stats}
  \vspace{-4pt}
  \setlength{\tabcolsep}{3pt}
  \begin{tabular}{lcccccc}
    \toprule
    \multirow{2}{*}{Model} & \multirow{2}{*}{Pr} &
      \multicolumn{2}{c}{Correct traces} &
      \multicolumn{2}{c}{Incorrect traces} \\ \cmidrule(lr){3-4}\cmidrule(lr){5-6}
      & & $\mu_1$ & $\sigma_1$ & $\mu_2$ & $\sigma_2$ \\
    \midrule
    R1-Distill-Qwen & 0.92 & 7.2 & 4.8 & 15.4 & 6.0 \\
    QwQ-32B         & 0.97 & 10.7 & 5.7 & 21.3 & 6.7 \\
    DeepSeek-V3     & 0.66 & 2.2 & 1.6 & 4.5  & 3.2 \\
    \bottomrule
  \end{tabular}
  \vspace{-8pt}
\end{wraptable}
\paragraph{Testing the “shorter-trace’’ hypothesis.}

Using Equation~\eqref{eq:1} with a neutral prior ($\alpha=0.5$), we estimate the probability that FFS returns a correct answer, for each model (distributions of text lengths shown in Figure~\ref{fig:hypothesis1_modelwise}). Table~\ref{tab:length-stats} reports this probability (Pr) together with normal parameters for the trace-length distributions of correct and incorrect generations. The two reasoning models (R1-Distill-Qwen and QwQ-32B) show high success probabilities (92\% and 97\%) since their correct traces are substantially shorter than their incorrect ones.  DeepSeek-V3, a non-reasoning model, exhibits a much smaller gap and a lower predicted success of 66\%, mirroring the limited empirical gains we observed for FFS on this model (Table~\ref{tab:tts-results-2-fancy}c).

\section{Conclusion}

We introduced FFS, a training-free and compute-efficient TTS strategy. FFS exploits the empirical observation that, in reasoning models, shorter traces are more likely to be correct: it launches multiple stochastic decodings in parallel and returns the first to finish. Our theoretical analysis and experiments on AIME24, AIME25, and GPQA Diamond with DeepSeek-R1, QwQ-32B, R1-Distill-Qwen, and Phi-4-Reasoning-Plus show that FFS matches or surpasses strong baselines like MV, BS, and BF, while reducing token usage by up to 45\%. We further demonstrated that increasing the number of parallel samples improves both accuracy and latency, giving FFS favourable scaling properties.

\paragraph{Limitations and future work.} FFS assumes that correct traces tend to terminate earlier than incorrect ones. This pattern holds for the reasoning-oriented models we study, but it is weaker for non-reasoning models such as DeepSeek-V3, where the separation between correct and incorrect trace lengths is small. In such settings, FFS yields limited gains and can be outperformed by alternatives like LFS. Future work could combine FFS with deeper or revisional decoding when model confidence is low, producing a hybrid system that adapts to task difficulty and mitigates the above limitation. Overall, our results illustrate how simple inference-time strategies can unlock large efficiency gains without additional training, and we hope this work inspires further research on lightweight test-time scaling for LLM reasoning.

\bibliography{references}
\bibliographystyle{abbrvnat}

\newpage

\section{Theoretical Results and Proofs}

\subsection{Short traces are more likely to be correct}
\label{reversehypothesis}




Let $C$ be the event that a randomly sampled trace is correct, while $S$ be the event that the trace is short, where we call a trace $T$ short if $|T| \leq s$, for some small length $s$. Let us call $L = \frac{\Pr(C \mid S)}{\Pr(C)}$ the ``relative lift'' of $C$ given $S$, or in other words, the factor by which the probability of $C$ increases if it becomes known that $S$ has occurred. We would like to show that if correct traces are more likely to be short then $L$ should be larger than 1.

\begin{align*}
\Pr(C \mid S) =  \frac{\Pr(S \mid C)\cdot \Pr(C)}{\Pr(S \mid C)\cdot \Pr(C) + \Pr(S \mid \neg C)\cdot \Pr(\neg C)}\\
\end{align*}

Or,

\begin{align*}
L = \frac{\Pr(C \mid S)}{\Pr(C)} &=  \frac{\Pr(S \mid C)}{\Pr(S \mid C)\cdot \Pr(C) + \Pr(S \mid \neg C)\cdot \Pr(\neg C)}\\
                                &= \frac{1}{\Pr(C) + \frac{\Pr(S \mid \neg C)}{\Pr(S \mid C)}\cdot \Pr(\neg C)}\\
\end{align*}

But we know $\frac{\Pr(S \mid \neg C)}{\Pr(S \mid C)}$ is vanishingly small as correct traces are much more likely to be short than incorrect ones. Hence we have,

\begin{align*}
L &=  \frac{1}{\Pr(C)}\\
\end{align*}

If $\Pr(C)$ is not yet saturated then we get a lift of $\frac{1}{\Pr(C)}$ which is greater than $1$. Notice that our assumption of correct traces being more likely to be short relies on a appropriate definition of ``shortness,'' and hence a suitable value of $s$. A good choice would be an $s$ which is larger than a significant proportion of correct trace lengths, while at the same time smaller than a major portion of incorrect trace lengths. 

\subsection{Proof of Result 1}
\label{result1proof}

Using Bayes’ theorem:
\[
\Pr\big(T \text{ is correct} \mid |T| = x\big) = \cfrac{\Pr\big(T \text{ is correct}, |T| = x\big)}{\Pr\big(T \text{ is correct}, |T| = x\big) + \Pr\big(T \text{ is incorrect}, |T| = x\big)}.
\]

Applying the definition of conditional probability:
\[
= \cfrac{\Pr\big(|T| = x \mid T \text{ is correct}\big) \cdot \Pr\big(T \text{ is correct}\big)}{\Pr\big(|T| = x \mid T \text{ is correct}\big)\cdot \Pr\big(T \text{ is correct}\big) + \Pr\big(|T| = x \mid T \text{ is incorrect}\big) \cdot \Pr\big(T \text{ is incorrect}\big)}.
\]

Letting the accuracy be $\alpha = \Pr[T \text{ is correct}]$, and thus $\Pr(T \text{ is incorrect}) = 1 - \alpha$, we write:
\[
= \cfrac{\alpha \cdot \mathrm{pdf}_{\mathrm{correct}}(x)}{\alpha \cdot \mathrm{pdf}_{\mathrm{correct}}(x)  + (1 - \alpha)\cdot \mathrm{pdf}_{\mathrm{incorrect}}(x)}.
\]

Assuming that $|T|$ is normally distributed for both correct and incorrect cases:
\[
\mathrm{pdf}_{\mathrm{correct}}(x) = \frac{1}{\sigma_1 \sqrt{2\pi}} \exp\left[-\frac{1}{2} \left(\frac{x - \mu_1}{\sigma_1}\right)^2\right].
\]

\[
\mathrm{pdf}_{\mathrm{incorrect}}(x) = \frac{1}{\sigma_2 \sqrt{2\pi}} \exp\left[-\frac{1}{2} \left(\frac{x - \mu_2}{\sigma_2}\right)^2\right].
\]

Substituting these into our expression:
\[
\Pr\big[T \text{ is correct} \mid |T| = x\big] = 
\frac{
\frac{1}{\sigma_1 \sqrt{2\pi}} \exp\left[-\frac{1}{2} \left(\frac{x - \mu_1}{\sigma_1}\right)^2\right] \cdot \alpha
}{
\frac{1}{\sigma_1 \sqrt{2\pi}} \exp\left[-\frac{1}{2} \left(\frac{x - \mu_1}{\sigma_1}\right)^2\right] \cdot \alpha +
\frac{1}{\sigma_2 \sqrt{2\pi}} \exp\left[-\frac{1}{2} \left(\frac{x - \mu_2}{\sigma_2}\right)^2\right] \cdot (1 - \alpha)
}.
\]

Canceling out the common $\sqrt{2\pi}$ and rearranging:
\[
\Pr\big(T \text{ is correct} \mid |T| = x\big) = 
\left(
1 + \frac{1 - \alpha}{\alpha} \cdot \frac{\sigma_1}{\sigma_2} \cdot 
\exp\left[-\frac{1}{2} \left[\left(\frac{x - \mu_2}{\sigma_2}\right)^2 - \left(\frac{x - \mu_1}{\sigma_1}\right)^2\right]\right]
\right)^{-1}.
\]

\subsection{Proof of Result 2}
\label{result2proof}

Suppose $Z_1, Z_2, \dots, Z_k$ are i.i.d standard normal RVs. Then from extreme value theory~\citep{galambos1977asymptotic} we have:

$$\lim_{k \rightarrow \infty} \frac{\mathbb{E}[\max\{Z_1, Z_2, \dots, Z_k\}]}{\sqrt{2\log k}} = 1.$$

Or, in other words, in the limit $n \rightarrow \infty$, we have

$$\mathbb{E}[\max\{Z_1, Z_2, \dots, Z_k\}] = \sqrt{2\log k}.$$

It following that with $Y_i \sim \mathcal{N}(\mu, \sigma)$ we have,

\begin{equation}
\boxed{\mathbb{E}\big[\max\{Y_1, Y_2, \dots, Z_k\}\big] = \mu + \sigma\sqrt{2\log k}}.
\end{equation}

For the expected minimum of the RVs, replace $Y_i$ with $-Y_i$ in the (3):

$$\mathbb{E}\big[\max\{-Y_1, -Y_2, \dots, -Y_k\}\big] = -\mu + \sigma\sqrt{2\log k},$$

which means

$$-\mathbb{E}\big[\min\{Y_1, Y_2, \dots, Y_k\}\big] = -\mu + \sigma\sqrt{2\log k},$$

and therefore

\begin{equation*}
\boxed{\mathbb{E}\big[\min\{Y_1, Y_2, \dots, Y_k\}\big] = \mu - \sigma\sqrt{2\log k}}.
\end{equation*}

\section{Experimental Details}
\label{appx:dataset}

\subsection{Reasoning templates}

Table~\ref{tab:aime_template} and~\ref{tab:gpqa_template} highlights the reasoning templates to generate the reasoning traces for AIME and GPQA examples, respectively. 


\begin{table}[h]
    \caption{Reasoning template for AIME24 and AIME25 tasks.}
    \vspace{0.5em}
    \centering
    \begin{adjustbox}{max width=0.95\textwidth}
    \begin{tcolorbox}[
        colback=templatebg,
        colframe=gray!40,
        boxrule=0.3pt,
        arc=2mm,
        left=3mm,
        right=3mm,
        top=2mm,
        bottom=2mm,
        fontupper=\sffamily\small
    ]
    \textbf{\textcolor{titlegray}{Task Template}} \\[0.5em]
    \textcolor{textblue}{\{problem\}} \\[0.75em]
    \textcolor{textgray}{Please reason step by step, and put your final answer within \texttt{\textbackslash boxed\{\}}.}
    \end{tcolorbox}
    \end{adjustbox}
    \label{tab:aime_template}
\end{table}


\begin{table}[H]
    \caption{Reasoning template for GPQA.}
    \vspace{0.5em}
    \centering
    \begin{adjustbox}{max width=0.95\textwidth}
    \begin{tcolorbox}[
        colback=templatebg,
        colframe=gray!40,
        boxrule=0.3pt,
        arc=2mm,
        left=3mm,
        right=3mm,
        top=2mm,
        bottom=2mm,
        fontupper=\sffamily\small
    ]
    \textbf{\textcolor{titlegray}{Task Template}} \\[0.5em]
    \textcolor{textgray}{What is the correct answer to this question:} \\
    \textcolor{textblue}{\{problem\}} \\[0.5em]
    \textcolor{textgray}{Choices:} \\
    \quad (A) \textcolor{textblue}{\{Option 1\}} \\
    \quad (B) \textcolor{textblue}{\{Option 2\}} \\
    \quad (C) \textcolor{textblue}{\{Option 3\}} \\
    \quad (D) \textcolor{textblue}{\{Option 4\}} \\[0.5em]
    \textcolor{textgray}{Answer: (A), (B), (C), or (D). Choose the correct option within \texttt{\textbackslash boxed\{\}}.}
    \end{tcolorbox}
    \end{adjustbox}
    \label{tab:gpqa_template}
\end{table}

\subsection{Output parsing logic}

We provide the Python implementation for parsing the final response from the reasoning traces for the AIME and GPQA datasets in Listings 1 and 2, respectively.

\vspace{1em}
\begin{lstlisting}[language=Python, caption=Parsing logic for traces for AIME24 and AIME25 examples.]
DIGITS = ["1", "2", "3", "4", "5", "6", "7", "8", "9", "0"]

import itertools
import re

def parse_output(output):
    last = output.rfind("\\boxed{")
    if last == -1:
        return "NO_ANSWER"
    cnt = 0
    start = last + len("\\boxed{")
    curr = start
    while(cnt >= 0 and curr < len(output)):
        if (output[curr] == "{"):
            cnt += 1
        elif (output[curr] == "}"):
            cnt -= 1
        curr += 1
    if cnt >= 0:
        return "NO_ANSWER"
    answer = output[start:curr-1]
    return answer

def convert_answer_to_aime_format(trace):
    answer = parse_output(trace)
    length = len(answer)
    char_is_digit = True
    for char in answer:
        if char not in DIGITS:
            char_is_digit = False
    
    if not char_is_digit or length > 3:
        return  "NO_ANSWER"
    return "0"*(3-length) + answer
\end{lstlisting}

\begin{lstlisting}[language=Python, caption=Parsing logic for traces for GPQA examples.]
pattern = r'(\\boxed\{.*?\})'
match = re.findall(pattern, text)
if len(match) > 0:
    return match[-1]
\end{lstlisting}

\subsection{Hyperparameters}
\label{hyperparams_section}

All hyperparameters for our experiments are given in Table~\ref{tab:hyperparams}.
\begin{table}[h]
  \centering
  \caption{Decoding hyper-parameters used in all experiments. 
           Values that are identical across datasets are shown once.}
    \vspace{0.5em}
  \label{tab:hyperparams}
  \begin{tabularx}{\linewidth}{lcccc}
    \toprule
    \textbf{Hyper-parameter}            & \textbf{GPQA Diamond} & \textbf{AIME24} & \textbf{AIME25-I} & \textbf{AIME25-II} \\
    \midrule
    Maximum generation tokens           & 16K & 32K & 32K & 32K \\
    Answer-reserve for BF   & 3K  & 3K  & 3K  & 3K  \\[2pt]
    \multicolumn{5}{c}{\emph{Global settings (shared by every dataset)}} \\[2pt]
    Top-$p$                             & \multicolumn{4}{c}{0.95} \\
    Temperature                         & \multicolumn{4}{c}{0.6}  \\
    Samples $n$ (MV/LFS/FFS)            & \multicolumn{4}{c}{4}    \\
    Beam width (Beam Search)            & \multicolumn{4}{c}{4}    \\
    \bottomrule
  \end{tabularx}
\end{table}

\section{Reasoning Trace Samples}
\label{appx:trace}

\textbf{Incorrect trace of DeepSeek-R1 on AIME24 problem 5 [longer]}

\tcbset{
  fonttitle=\bfseries,
  arc=1mm,
  boxrule=1pt,
  coltitle=black
}

\newtcolorbox{problem}[1][]{
  title=Problem,
  colback=blue!3!white,
  colframe=blue!20!white,
  #1
}

\newtcolorbox{trace}[1][]{
  title=Thinking,
  colback=gray!3!white,
  colframe=gray!40!white,
  fonttitle=\bfseries,
  arc=1mm,
  breakable,
  enhanced,
  #1
}

\newtcolorbox{answer}[1][]{
  title=Final Answer,
  colback=green!3!white,
  colframe=green!20!white,
  breakable,
  enhanced,
  #1
}

\begin{problem}

Let $p$ be the least prime number for which there exists a positive integer $n$ such that $n^{4}+1$ is divisible by $p^{2}$. Find the least positive integer $m$ such that $m^{4}+1$ is divisible by $p^{2}$. Please reason step by step, and put your final answer within $\boxed{}$.

\end{problem}

\begin{trace}
Okay, so I need to find the least prime number $p$ such that there's a positive integer n where $n^4 + 1$ is divisible by $p^2$. Then, find the smallest $m$ where $m^4 + 1$ is divisible by $p^2$. Hmm, let me start by understanding the problem step by step. 

First, primes $p$ where $n^4 \equiv -1 \mod p$. So, $n^4 \equiv -1 \mod p$. That equation suggests that the order of $n$ modulo $p$ divides $8$, because $n^8 \equiv 1 \mod p$. Since $n^4 \equiv -1 \mod p$, squaring both sides gives $n^8 \equiv 1 \mod p$, so the order is $8$. Therefore, $8$ divides $p-1$ because the multiplicative group modulo $p$ is cyclic of order $p-1$. Hence, $p \equiv 1 \mod 8$. Wait, but is that the only possibility? 

Wait, if $p$ is an odd prime, then the multiplicative group $\mod p$ has order $p-1$. So, for there to be an element of order $8$, $8$ must divide $p-1$, so $p \equiv 1 \mod 8$. So primes $p \equiv 1 \mod 8$. But the problem says ``the least prime $p$''. The primes congruent to $1 \mod 8$ start at $17$, $97$, etc. Wait, no, $17$ is $1 \mod 8$? Let me check: $17$ divided by $8$ is $2*8=16$, so remainder $1$. Yes, $17 \equiv 1 \mod 8$. Then $97$ is $12*8 +1=97$. But hold on, maybe there are smaller primes where $n^4 \equiv -1 \mod p$? Wait, but according to the logic above, if $n^4 \equiv -1 \mod p$, then n has order $8 \mod p$, so $8$ divides $p-1$, so $p \equiv 1 \mod 8$. Therefore, primes $p \equiv 1 \mod 8$. The smallest such primes are $17, 97, 113, 193,$ etc. Wait, but $17$ is the first prime $\equiv 1 \mod 8$. Then $97$ is next. 

But wait, the problem is not just asking for primes where $n^4\equiv -1 \mod p$, but where $p^2$ divides $n^4 +1$. So, primes $p$ where $p^2$ divides $n^4 +1$. So, we need to find primes $p$ such that $n^4 \equiv -1 \mod p^2$. Then, the least such prime $p$, and then the least $m$ for that $p$.

I remember that sometimes primes that satisfy a congruence modulo $p$ can be lifted to modulo $p^2$ using Hensel's lemma, but only if certain conditions are met. For example, if $f(n) \equiv 0 \mod p$ and $f'(n) \not\equiv 0 \mod p$, then there is a unique lift to $\mod p^2$. 

So, if we have $n^4\equiv -1 \mod p$, and the derivative $4n^3\not\equiv 0 \mod p$, then Hensel's lemma says we can lift $n$ to a solution modulo $p^2$. So, if $p \neq 2$, and $n$ is not $0 \mod p$, which it can't be since $n^4\equiv -1 \mod p$. So, $4n^3\equiv 0 \mod p$ would require $p$ divides $4n^3$, but since $p$ is prime and $n$ is not $0 \mod p$, $p$ must divide $4$. So, $p=2$. But if $p$ is $2$, then $n^4 +1$. Let's check $p=2$. Let's see $n=1: 1+1=2$, which is divisible by $2$, but not $4$. $n=3: 81 +1=82$, which is $2*41$, not divisible by $4$. So, $2^2=4$ does not divide $n^4 +1$ for any n. Because n odd: $n^4\equiv 1 \mod 4$, so $1+1=2 \mod 4$. So 2 isn't going to work. So, for primes $p\equiv 1 \mod 8$, we can have solutions to $n^4\equiv -1 \mod p$, and if the derivative $4n^3$ is not $0 \mod p$, which since $p\equiv 1 \mod 8$, $p$ is at least $17$, so $p$ doesn't divide $4$, and since $n \not\equiv 0 \mod p$, derivative non-zero. Therefore, Hensel's lemma applies, so solutions can be lifted to $\mod p^2$. 

Therefore, primes $p\equiv 1 \mod 8$ will have solutions $\mod p^2$. So, the smallest such prime is $17$. 

Wait, but before assuming that, maybe check smaller primes. For example, $p=5$. Let's see if there's an $n$ with $n^4\equiv -1 \mod 5$. Let's compute $n^4 \mod 5$. Since $1^4=1, 2^4=16\equiv 1, 3^4=81\equiv 1, 4^4=256\equiv 1$. So $n^4\equiv 1 \mod 5$ for $n$ not divisible by $5$. Therefore, $-1 \equiv 4 \mod 5$, but $n^4\equiv 1 \mod 5$. So no solutions. Similarly, $p=13$. Let's check $n^4 \mod 13$. Let's compute $1^4=1, 2^4=16\equiv 3, 3^4=81\equiv 3, 4^4=256\equiv 9, 5^4=625\equiv 625-48*13= 625-624 = 1, 6^4=1296\equiv 1296-99*13=1296-1287=9$, etc. So $n^4 \mod 13$ is $1,3,9$. So $-1 \mod 13$ is $12$. So no $n$ with $n^4\equiv 12 \mod 13$. So, $p=13$ is out. 

Similarly, $p=17$. Let's check if there is an n with $n^4\equiv-1 \mod 17$. $-1 \mod 17$ is $16$. So, $n^4\equiv 16 \mod 17$. Let's see. Let's compute $n^4$ for $n=1$ to $16$. 

$1^4=1$ 

$2^4=16$. 

Oh, so $2^4=16\equiv-1 \mod 17$. So $n=2$ is a solution. So $p=17$ is a prime where $n=2$, $2^4 +1=17$, which is divisible by $17$. Now, can we check if $17^2$ divides $2^4 +1$? $2^4 +1=17$, which is not divisible by $17^2=289$. So, $2^4 +1=17$. So, we need to find some $n$ where $n^4 +1$ is divisible by $17^2$. So, $n^4\equiv -1 \mod 17^2$. Since $n=2$ is a solution $\mod 17$, and the derivative $4n^3=4*8=32\equiv 15 \mod 17$, which is not $0 \mod 17$. So, Hensel's lemma tells us that there is a unique lift $\mod 17^2$. So, there exists some $n \equiv 2 \mod 17$ such that $n^4\equiv -1 \mod 17^2$. So, the minimal solution m would be $2 + 17k$ for some k. Let's compute $(2 +17k)^4 \mod 17^2$. 

First, compute $(2 +17k)^4$. Let's expand using the binomial theorem: 

$= 2^4 + 4*2^3*(17k) + 6*2^2*(17k)^2 + 4*2*(17k)^3 + (17k)^4$ 

Modulo $17^2$, terms with $(17k)^2$ or higher will be multiples of $17^2$, so they can be ignored. Therefore, modulo $17^2$, this is: 

$2^4 + 4*2^4*(17k) \mod 17^2$ 

Compute $2^4=16$. 

Then, $4*8*17k=32*17k$. $32 \equiv 15$ $\mod 17$, so $15*17k=255k$. But $255=15*17$, so $255k=15*17k$. Thus, $\mod 17^2$, $255k\equiv 15*17k \mod 17^2$. So, the entire expression is $16 +15*17k \mod 17^2$. 

We need this to be $\equiv -1 \mod 17^2$. Since $-1 \mod 17^2$ is $289-1=288$. So, $16 +15*17k \equiv 288 \mod289$. 

Subtract 16: $15*17k \equiv 272 \mod 289$. 

Compute $15*17=255$. So, $255k \equiv 272 \mod 289$. 

Solve $255k \equiv 272 \mod 289$. 

First, note that $255$ and $289$: $289$ is $17^2$, and $255=15*17$. So $\gcd(255,289)=17$. Let's check if $17$ divides $272$. $272$ divided by $17$ is $16$. So, $17*16=272$. So, the equation is $15*17k \equiv 17*16 \mod 17^2$. 

Divide both sides by $17$: $15k \equiv 16 \mod 17$. 

Solve $15k \equiv 16 \mod 17. 15^-1 \mod 17$. $15*?\equiv 1 \mod 17$. $15*15=225\equiv 225-13*17=225-221=4 \mod 17$. $15*4=60\equiv60-3*17=60-51=9  \mod 17$. $15*7=105\equiv 105-6*17=105-102=3 \mod 17. 15*12=180\equiv 180-10*17=180-170=10 \mod 17$. Wait, maybe better way: $17=15+2, 15=2*7+1, $so $ 1=15-2*7$ $=15 - (17-15)*7$ $=15 -7*17 +7*15$ $=8*15 -7*17$. So inverse of $15 \mod 17$ is $8$. Therefore, $15*8=120$ $\equiv 120-7*17=120-119$ $=1 \mod 17$. So inverse of $15$ is $8$. Therefore, $k\equiv 16*8=128\equiv 128-7*17=128-119=9 \mod 17$. 

Therefore, $k\equiv 9 \mod 17$. So the minimal positive k is $9$. Therefore, $n=2 +17*9$ $=2+153=155$. 

Therefore, $m=155$. But wait, let's check if $155^4 +1$ is divisible by $17^2=289$. 

But $155 \mod 17$: $17*9=153$, so $155 \equiv 2 \mod 17$. Then $2^4 +1=17$, which is $0 \mod 17$, as expected. Now, to check $\mod 289$. Let's compute $155^4 +1 \mod 289$. 

But calculating $155^4$ is tedious. Maybe there's a smarter way. Since we already used Hensel's lemma, the lift is unique, so $155$ should work. But just to confirm, perhaps compute $(2 +17*9)^4 +1 \mod 289$. 

Let me compute $(2 +17*9)^4 = (2 +153)^4 =155^4$. Let's compute $155 \mod 289$ is $155$. 

First, compute $155^2=24025$. Then $24025 \mod 289$. Let's compute how many times $289$ goes into $24025$. 

$289*80=23120$. $24025-23120=905$. $289*3=867$. $905-867=38$. So, $155^2$ $\equiv 38 \mod 289$. 

Then, compute $38^2$ $=1444$. $1444 \mod 289$. $289*5=1445$. So $1444=1445-1$ $\equiv-1 \mod 289$. Therefore, $155^4$ $=(155^2)^2 \equiv(-1)^2$ $=1 \mod 289$. Therefore, $155^4 +1\equiv 1+1=2 \mod 289$. Wait, that's not $0$. That contradicts our earlier conclusion. Hmm, so something's wrong here. Where did I go wrong? 

Wait, let's retrace. So, using Hensel's lemma, the solution should exist. So why is $155^4 +1 \equiv 2 \mod289$? That suggests an error in the calculation. 

Wait, first, let me compute $155^2$. $155*155$. Let's compute $150*150=22500$, $150*5=750, 5*150=750$, $5*5=25$. So, $(150+5)^2=150^2 +2*150*5 +5^2=22500 +1500 +25=24025$. Correct. Then $24025$ divided by $289$. Let's compute $289*80$ $=23120$, as before. $24025-23120$ $=9054$. Then, $289*3$ $=867$. $905-867$ $=38$. So $155^2$ $\equiv38 \mod289$. Then $38^2=1444$. $289*5=1445$. $1444=1445-1\equiv-1 \mod289$. Therefore, $155^4=(155^2)^2\equiv(-1)^2=1 \mod289$. Therefore, $155^4 +1\equiv 1+1=2$ $\equiv2 \mod289$. So $2\equiv2 \mod289$, not $0$. So this is a problem. That suggests that my earlier reasoning was wrong. 

Hmm, so why? Maybe I made a mistake in Hensel's lemma application. Let's recall Hensel's lemma. If $f(n)\equiv0 \mod p$ and $f'(n)\not\equiv0 \mod p$, then there exists a unique lift $\mod p^2$. Here, $f(n)=n^4 +1$. Then $f'(n)=4n^3$. So, for $n=2 \mod 17, f'(2)=4*8=32\equiv 15 \mod 17\neq0$. So, Hensel's lemma says there is a unique solution $\mod 17^2$ lifting $n\equiv2 \mod 17$. So, why does $155$ not work? 

Wait, maybe my calculation is wrong. Let me check $155^4 +1$ again. Alternatively, compute $(2 +17k)^4 +1 \mod289$. 

We had earlier that $(2 +17k)^4 \equiv 16 +15*17k \mod289$. So, setting that equal to$0 \mod289$, we get $16 +15*17k \equiv0 \mod289$. 

So $15*17k\equiv-16 \mod289$. Which is $ 255k\equiv273 \mod289$. Wait, earlier I thought it was $272$, but $-16 \mod289$ is $273$. Wait, yes, $289-16=273$. So $255k\equiv273 \mod289$. Then, as before, divide both sides by $17$: $15k\equiv273/17 \mod 17$. $273$ divided by $17$ is $16.058$... Wait, $17*16=272$, so $273=17*16 +1$. So $273=17*16 +1$. Therefore, $273/17=16 +1/17$. So, $255k\equiv273 \mod289$ is equivalent to $15k\equiv 16 +1/17 \mod 17$. Wait, but dividing both sides by $17$ in the congruence $255k\equiv273 \mod289$, which is $17*15k\equiv 17*16 +1 \mod 17^2$. Therefore, divide both sides by $17$: $15k\equiv 16 +1/17 \mod 17$. Wait, but $1/17 \mod 17$? That doesn't make sense. Hmm, perhaps I made a mistake here. 

Alternatively, maybe there's a miscalculation in the earlier steps. Let's start over. 

We have $n \equiv2 \mod 17$, and we want to find $k$ such that $(2 +17k)^4 \equiv-1 \mod289$. 

Compute $(2 +17k)^4$: 

$= 2^4 + 4*2^3*(17k) + 6*2^2*(17k)^2 + 4*2*(17k)^3 + (17k)^4$ 

Modulo $289$, terms with $(17k)^2, (17k)^3, (17k)^4$ are multiples of $289$, so they vanish. So, we have: 

$2^4 + 4*2^3*(17k) \mod289 =16 + 4*8*17k$ 

$=16 + 32*17k$ 

$32*17=544$.  

So, $16 +544k \mod289$. 

But $544 \mod289=544-2*289=544-578= -34$. So, $544k \equiv-34k \mod289$. 

Thus, the expression is $16 -34k \mod289$. 

Set this equal to $-1 \mod289$: 

 $16-34k \equiv-1 \mod289$ 

So, $-34k \equiv-17 \mod289$ 

Multiply both sides by $-1$: $34k \equiv 17 \mod289$ 

Divide both sides by $17: 2k \equiv 1 \mod 17$ 

So, $2k \equiv 1 \mod 17$. Multiply both sides by inverse of $2 \mod 17$, which is $9$ because $2*9=18\equiv 1 \mod 17$. So, $k\equiv9 \mod 17$. 

Thus, $ k=9 +17t$. The minimal positive k is $9$. Thus, $n=2 +17*9=155$. Therefore, $m=155$. But according to our previous calculation, $155^4 +1\equiv2 \mod289$. Which contradicts. Therefore, my error must be in the calculation of $155^4 +1 \mod289$. 

Wait, let me compute $155^2$ again. $155^2=24025$. $24025$ divided by $289$:

$289*80=23120$

$24025-23120=905$

$289*3=867$

$905-867=38$

So $155^2\equiv38 \mod289$. Then, $38^2=1444$. $1444$ divided by $289$:

$289*5=1445$

$1444=1445 -1$, so $38^2\equiv-1 \mod289$. Therefore, $155^4=(38)^2\equiv(-1) \mod289$? Wait, no: $38^2\equiv-1 \mod289$. Therefore, $155^4\equiv(-1) \mod289$. Then $155^4 +1\equiv-1 +1$ $=0 \mod289$. Wait, but then why did I think before that $38^2=1444$ $\equiv-1 \mod289$? Let's check $1444 \mod289$. 

$289*5=1445$. So $1444=1445-1$. Therefore, $1444\equiv$ $-1 \mod289$. Therefore, $38^2\equiv-1 \mod289$, so $155^4=(38)^2\equiv(-1) \mod289$. Therefore, $155^4 +1\equiv-1 +1$ $=0 \mod289$. So, $155^4 +1$ is divisible by $289$. Wait, then my previous calculation was wrong. How did I get $155^4 +1\equiv2 \mod289$ before? Maybe miscalculation. Let me check step by step.

$155^2=24025$. $24025$ divided by $289$: $289*80=23120,$ $ 24025-23120=905$. $905$ divided by $289$: $289*3=867$, $905-867=38$. So $155^2\equiv38 \mod289$. Then $38^2=1444$. $1444$ divided by $289: 289*5=1445$, so $1444=1445-1\equiv-1 \mod289$. So $155^4=(155^2)^2\equiv(-1) \mod289$. Therefore, $155^4 +1\equiv-1 +1=0 \mod289$. Therefore, $155^4 +1$ is divisible by $289$. So my earlier assertion that $155^4 +1\equiv2 \mod289$ was incorrect. Probably a miscalculation. Therefore, $m=155$ is indeed the minimal solution.

But wait, the problem asks for the least positive integer $m$ such that $m^4 +1$ is divisible by $p^2=17^2=289$. So, if $155$ is the minimal solution, but maybe there is a smaller $m$. Since we have to find the minimal $m$, let's check numbers congruent to $2 \mod 17$ less than $155$. Because from Hensel's lemma, the solution is unique $\mod 17^2$, so the minimal positive solution is $155$. But $155=2 +17*9$. Is there a smaller positive integer congruent to $2 \mod 17$ that works? For example, $2 +17*1=19, 2 +17*2=36, \dots,$ up to $155$.

Wait, but Hensel's lemma gives a unique solution $\mod 17^2$, so the minimal positive solution is $155$, but let's verify with $m=19$. Compute$19^4 +1$. $19^2=361$. $361^2=130321$. $130321 +1$ $=130322$. Now divide by $289$: $289*450=130050$. $130322 -130050=272$. $272$ is divisible by $17^2$? $17^2=289$, which is bigger than $272$. So no. So, $19^4 +1\equiv272 \mod289\neq0$. Similarly, check $m=36$. $36^2=1296, $ $36^4=1296^2=1679616$. $1679616 +1=1679617$. Divide by $289$: $289*5800$  $=1676200$. $1679617$ $ -1676200=3417$. $3417$ divided by $289$: $289*11=3179$, $3417-3179=238$. $238$ is less than $289$. So $36^4 +1$ $\equiv238 \mod289\neq0$. So, not divisible. 

Similarly, check $m=53$. $53^2=2809$, $53^4=2809^2=7890481$. $7890481 +1=7890482$. Divide by $289$: $289*27300$ $=78900*289=$Wait, maybe better to compute $\mod289$ directly. $53 \mod289$ is $53$. $53^2=2809$. $2809 \mod289$: $289*9=2601$, $2809-2601=208$. $208^2=43264$. $43264 \mod289$. $289*149=289*(150-1)$ $=289*150 -289=$ $43350 -289=43061$. $43264 -43061=203$. So $53^4 +1\equiv203 +1=$ $204 \mod289\neq0$. So, not divisible. 

Similarly, $m=70$. $70 \mod289$ is $70$. $70^2=4900$. $4900 \mod289$: $289*16=4624$, $4900-4624=276$. $276^2=76176$. $76176 \mod289$. $289*263=289*(260 +3)$ $=289*260 +289*3$ $=75140 +867=76007$. $76176 -76007=169$. So $70^4 +1\equiv 169 +1=170 \mod289\neq0$. 

Continue up to $m=155$. Since all smaller $m\equiv2 \mod 17$ don't work, and Hensel's lemma tells us $155$ is the unique solution $\mod289$, so $155$ is the minimal solution. Therefore, the answer is $m=155$. 

But let me check another approach. Maybe the minimal m is $2$, but $2^4 +1=17$, not $289$. $38$: wait, $38^4 +1$. Wait, $38^2=1444$, $1444^2=2085136$. $2085136 +1=2085137$. Divide by $289: 289*7214=289*(7000 +214)=289*7000 +289*214=2023000 +61846=2084846$. $2085137 -2084846=291$. $291 \mod289=2$. So not divisible. 

Alternatively, since Hensel's lemma gives a unique solution, the minimal $m$ is $155$. Therefore, the answer is $155$. 

But let me check in the literature. I recall that $2$ is the smallest solution $\mod 17$, and lifting it gives$155$. Also, since$17$ is the smallest prime$\equiv 1 \mod8$ where $n^4 +1\equiv0 \mod p^2$. Wait, but wait, maybe $p=2$? But $2^2=4$ divides $n^4 +1$? If $n$ is odd, $n^4\equiv 1 \mod4$, so $n^4 +1\equiv2 \mod4$, so not divisible by $4$. So $p=2$ is invalid. Next prime, $p=17$. So, the minimal prime $p$ is $17$, and minimal $m$ is $155$. 

Yes, I think that's correct. Therefore, the answer is $\boxed{155}$.
\end{trace}
\begin{answer}

To find the least prime \( p \) such that there exists a positive integer \( n \) where \( n^4 + 1 \) is divisible by \( p^2 \), we start by noting that \( n^4 \equiv -1 \pmod{p} \). This implies that the order of \( n \) modulo \( p \) is 8, meaning \( 8 \) must divide \( p-1 \). Therefore, \( p \equiv 1 \pmod{8} \). The smallest prime \( p \) satisfying this is 17.

Next, we need to find the smallest positive integer \( m \) such that \( m^4 + 1 \) is divisible by \( 17^2 = 289 \). Using Hensel's lemma, we lift the solution from modulo 17 to modulo 289. We start with \( n = 2 \) since \( 2^4 \equiv -1 \pmod{17} \).

We expand \( (2 + 17k)^4 \) using the binomial theorem and consider terms modulo 289. The relevant terms are:
\[
(2 + 17k)^4 \equiv 2^4 + 4 \cdot 2^3 \cdot 17k \pmod{289}
\]
\[
= 16 + 32 \cdot 17k
\]
\[
= 16 + 544k
\]
Since $544$ modulo $289$ is $-34$, we have:
\[
16 - 34k \equiv -1 \pmod{289}
\]
Solving for \( k \):
\[
-34k \equiv -17 \pmod{289}
\]
\[
34k \equiv 17 \pmod{289}
\]
\[
2k \equiv 1 \pmod{17}
\]
The multiplicative inverse of 2 modulo 17 is 9, so:
\[
k \equiv 9 \pmod{17}
\]
The smallest positive \( k \) is 9, giving \( m = 2 + 17 \cdot 9 = 155 \).

Verification shows \( 155^4 + 1 \equiv 0 \pmod{289} \), confirming \( m = 155 \).

Thus, the least positive integer \( m \) such that \( m^4 + 1 \) is divisible by \( 17^2 \) is \(\boxed{155}\).

\end{answer}

\textbf{Correct trace of DeepSeek-R1 on AIME24 problem 5 [shorter]}

\begin{problem}
Let $p$ be the least prime number for which there exists a positive integer $n$ such that $n^{4}+1$ is divisible by $p^{2}$. Find the least positive integer $m$ such that $m^{4}+1$ is divisible by $p^{2}$. Please reason step by step, and put your final answer within $\boxed{}$.
\end{problem}

\begin{trace}
    
Okay, so I need to find the least prime number $p$ such that there's a positive integer $n$ where $n^4 + 1$ is divisible by $p$ squared. Then, I have to find the smallest positive integer $m$ where $m^4 + 1$ is divisible by that $p$ squared. Hmm, this seems like a number theory problem involving primes and modular arithmetic. Let me think step by step.

First, let me recall that if $n^4 \equiv -1 \mod p^2$, then $n^8 \equiv 1 \mod p^2$. So the order of $n$ modulo $p^2$ divides $8$. But since $n^4 \equiv -1 \mod p^2$, the order can't be $1$, $2$, or $4$ because squaring $-1$ gives $1$, so maybe the order is $8$. Therefore, $8$ divides the order of the multiplicative group modulo $p^2$. The multiplicative group modulo $p^2$ has order $p(p - 1)$. So $8$ divides $p(p - 1)$. Since $p$ is prime, either $p = 2$ or $p \equiv 1 \mod 8$. Wait, but $2$ is a prime. Let me check $p=2$ first.

If $p=2$, then $p^2=4$. So we need $n^4 +1 \equiv 0 \mod 4$. Let's check $n=1$: $1+1=2\equiv 2 \mod 4$. Not $0$. $n=2$: $16+1=17\equiv1 \mod 4$. $n=3$:$81+1=82\equiv2 \mod4$. So no $n$ exists such that $n^4 +1 \equiv0 \mod4$. Therefore, $p=2$ is invalid. So $p$ must be $1 \mod8$. The primes congruent to $1 \mod8$ are $17$, $97$, etc. Wait, but maybe smaller primes. Wait, $17$ is the next prime after $2$, $3$, $5$, $7$, $11$, $13$. Let me check $p=5$. Wait, $5$ is $5 \mod 8$, right? $5 \mod 8$ is $5$. Similarly, $17$ is $1 \mod 8$.

Wait, maybe I need to check primes where $p$ divides $n^4 +1$. So maybe primes where $-1$ is a fourth power residue modulo $p$. For $p \equiv1 \mod 8$, since the multiplicative group is cyclic of order $p-1$, which is divisible by $8$, so there are elements of order $8$, hence $x^4 = -1$ would have solutions. So primes $p \equiv1 \mod 8$ might be candidates. But maybe also other primes where $p\equiv5 \mod 8$? Let me think.

Wait, for primes $p \equiv1 \mod 4$, $-1$ is a quadratic residue. For primes $p\equiv1 \mod 8$, it's also a quartic residue. Hmm. So maybe the primes where $-1$ is a fourth power residue are primes $p\equiv1 \mod 8$. Let me check that. Suppose $p\equiv1 \mod 8$. Then the multiplicative group $\mod p$ has order $p-1 \equiv0 \mod 8$, so there exists an element of order $8$. Let $g$ be such an element. Then $g^4$ would have order $2$, so $g^4 \equiv -1 \mod p$. Therefore, $x^4 \equiv -1 \mod p$ has solutions. So such primes $p\equiv1 \mod 8$ would divide $n^4 +1$ for some $n$.

But the problem is asking for primes where $p^2$ divides $n^4 +1$. So I need primes $p$ where the equation $x^4 \equiv -1 \mod p^2$ has solutions. So first, $p$ must divide $n^4 +1$, so $p$ is as above, $p\equiv1 \mod 8$. Then, lifting solutions from $\mod p$ to $\mod p^2$. By Hensel's lemma, if $f(n)=n^4 +1$, and if $f(n)\equiv0 \mod p$ and $f'(n)\equiv0 \mod p$, then there's a unique solution $\mod p^2$. Wait, but Hensel's lemma says that if $f(n)\equiv0 \mod p$ and $f'(n)\not\equiv 0 \mod p$, then there's a unique lift. If $f'(n)\equiv0 \mod p$, then maybe multiple solutions or no solutions.

So first, let me suppose that $p\equiv1 \mod 8$. Let’s pick the smallest such prime, which is $17$. Then check if there exists $n$ such that $n^4 \equiv -1 \mod17^2$. Alternatively, maybe check $97$, but $17$ is smaller.

Wait, maybe even $5$? Let me check. For $p=5$, does there exist $n$ with $n^4 \equiv -1 \mod 25$? Let's compute $n^4 \mod 25$. Since $5\equiv5 \mod 8$, maybe $n^4 \equiv -1 \mod5 $is possible?

Wait, for $p=5$. Let's check $n^4 \mod5$. The fourth powers$ \mod5$: $0^4=0$, $1^4=1$, $2^4=16\equiv1$, $3^4=81\equiv1$, $4^4=256\equiv1$. So $n^4 \equiv0$ or$1 \mod5$. Therefore, $n^4 \equiv-1 \mod5$ is equivalent to $1\equiv-1 \mod5$, which is $2\equiv0 \mod5$. Not possible. So $p=5$ is out.

Similarly, check $p=13$. $13\equiv5 \mod 8$. So again, $n^4 \mod13$: Let's compute. $1^4=1$, $2^4=16\equiv3$, $3^4=81\equiv3$, $4^4=256\equiv9$, $5^4=625\equiv625-48*13$=$625-624=1$, $6^4=1296\equiv1296-99*13=1296-1287=9$, $7^4=2401\equiv2401-184*13=2401-2392=9$, $8^4=4096\equiv4096-315*13=4096-4095=1$, etc. So possible residues are $1$,$3$,$9$. So $-1 \mod13$ is $12$, which is not among them. Hence $p=13$ is out.

Similarly, $p=17$, which is $1 \mod 8$. Let's check if there exists n with $n^4 \equiv-1 \mod17$. Let's compute $4$th powers $\mod17$.

Let’s list $n$ from $1$ to $16$:

$1^4 =1$

$2^4=16$

$3^4=81\equiv81-4*17=81-68=13$

$4^4=256\equiv256-15*17=256-255=1$

$5^4=625\equiv625-36*17=625-612=13$

$6^4=1296\equiv1296-76*17=1296-1292=4$

$7^4=2401\equiv2401-141*17=2401-2397=4$

$8^4=4096\equiv4096-241*17=4096-4097=-1\equiv16 \mod17$

Wait, $8^4\equiv-1 \mod17$. So $n=8$ satisfies $n^4\equiv-1 \mod17$. Therefore, $p=17$ is a candidate. Now, we need to check if there exists n such that $n^4\equiv-1 \mod17^2$. If so, then $p=17$ is the minimal prime. Since $17$ is the smallest prime $\equiv1 \mod 8$, and since smaller primes like $2$,$5$,$13$ don't work, maybe $17$ is the answer. Then, we need to find the minimal m such that$ m^4\equiv-1 \mod 289 (17^2)$.

So first, let's confirm that $17$ is indeed the minimal prime. Next prime after $17$ is $97$, which is larger, so $17$ is the candidate.

So to confirm that $17$ is the least prime, we have to check all primes less than $17$. The primes less than $17$ are $2$,$3$,$5$,$7$,$11$,$13$. For these primes, as checked above, none of them have $n^4 \equiv-1 \mod p $. Except $p=17$, which does. So $17$ is the least such prime.

Now, need to find the minimal $m$ such that$ m^4 \equiv -1 \mod289$. Let's use Hensel's lemma. We know that $8^4 \equiv-1 \mod17$. Let's try lifting this solution to $\mod17^2$.

Let $m =8 +17k$, where $k$ is an integer. Then compute $m^4 +1 \mod17^2$.

First, expand $(8 +17k)^4$.

Using binomial theorem:

$(8 +17k)^4 =8^4 +4*8^3*(17k) +6*8^2*(17k)^2 +4*8*(17k)^3 + (17k)^4$.

Compute each term modulo $17^2=289.$

First term:$8^4=4096$. We know $4096 \mod289$. Let me compute $289*14=4046$. $4096-4046=50$. So $4096\equiv50 \mod289$. Then $4096 +1=50+1=51\equiv51 \mod289$. So we need the sum of all terms to be $\equiv0 \mod289$. So:

$(8 +17k)^4 +1 \equiv50 +1 +4*8^3*17k +6*8^2*(17k)^2 +\dots \mod289$.

But let me compute each term:

First term: $8^4=4096\equiv50 \mod289$.

Second term:$4*8^3*17k=4*512*17k=2048*17k$. $2048 \mod289$: $289*7=2023$, $2048-2023=25$. So $25*17k=425k$. $425 \mod289=425-289=136$. So$ 136k \mod289$.

Third term:$6*8^2*(17k)^2=6*64*289k^2$. Since $289k^2\equiv0 \mod289$, this term is$ 0 \mod289$.

Fourth term:$4*8*(17k)^3=32*4913k^3$. $4913=17^3$, so $4913\equiv0 \mod289$. Hence this term is $0 \mod289$.

Fifth term:$(17k)^4=17^4k^4=83521k^4\equiv0 \mod289$.

So, combining all terms:

$(8 +17k)^4 +1 \equiv50 +1 +136k \mod289 \equiv51 +136k \mod289$.

We need this $\equiv0 \mod289$, so:

$51 +136k \equiv0 \mod289 \implies 136k \equiv-51 \mod289$.

Let me compute $-51 \mod289$: $289 -51=238$. So $136k\equiv238 \mod289$.

We need to solve for k: $136k\equiv238 \mod289$.

First, note that $\gcd(136,289)$. $289=17^2$. $136=17*8$. So $\gcd(136,289)=17$. Since $17$ divides $136$ and $289$. So the equation $136k\equiv238 \mod289$ has solutions iff $17$ divides $238$. Let's check:$238/17=14$. So $17*14=238$. Therefore, $17$ divides $238$, so solutions exist.

The equation reduces to $(136/17)k \equiv(238/17) mod(289/17)$: $8k\equiv14 \mod17$.

So $8k\equiv14 \mod17$. Multiply both sides by inverse of $8 \mod17$. $8*2=16\equiv-1$, so $8*15=120\equiv120-7*17=120-119=1 \mod17$. So inverse of $8$ is $15 \mod17$.

Thus, $k\equiv14*15 \mod17$. $14*15=210\equiv210-12*17=210-204=6 \mod17$. So $k\equiv6 \mod17$. So the solutions are $k=6 +17t$. Therefore, the minimal positive $k$ is $6$.

Therefore, $m=8 +17*6=8 +102=110$. Therefore, $m=110$ is the minimal positive integer such that $m^4 +1\equiv0 \mod289$.

But wait, let me verify this. Let's compute $110^4 +1 \mod289$.

But $110 \mod17$: $17*6=102$, $110-102=8$. So $110\equiv8 \mod17$. Then $8^4 +1=4096 +1=4097$. $4097$ divided by $17$: $17*241=4097$. Therefore, $4097=17*241$. Then $241$ divided by $17$:$17*14=238$, $241-238=3$. So $241=17*14 +3$. So $4097=17^2*14 +17*3$. So $4097=289*14 +51$. Therefore, $4097 \mod289=51$. So $110^4 +1\equiv51 \mod289$. But we needed it to be $0$. Wait, that contradicts. So what did I do wrong?

Wait, no. Wait, $m=110$. Let me compute $110^4 +1$. Alternatively, perhaps my calculation in Hensel's lemma was wrong. Let's go back.

Wait, when I lifted the solution from $\mod 17$ to $\mod289$, I had $m=8 +17k$. Then expanding $(8 +17k)^4$:

I calculated up to the linear term. The first term was $8^4=4096\equiv50 \mod289$. Then the linear term was $4*8^3*17k=4*512*17k=2048*17k$. But $2048 \mod289$: $289*7=2023$, $2048-2023=25$. So $25*17k=425k$. Then $425 \mod289=425-289=136$. So $136k \mod289$. Then total sum was $50 +1 +136k$. Wait, but $8^4 +1=4096 +1=4097\equiv4097 \mod289$. $4097$ divided by $289$: $289*14=4046$, $4097-4046=51$. So $8^4 +1\equiv51 \mod 289$. Then the linear term is $136k$. So $(8 +17k)^4 +1 \equiv51 +136k \mod289$. So to get this $\equiv0 \mod289$, $136k\equiv-51 \mod289$. But $-51 \mod289$ is $238$. So $136k\equiv238 \mod289$.

As $\gcd(136,289)=17$, which divides $238$, so solutions exist. Dividing equation by $17$: $8k\equiv14 \mod17$. Then inverse of $8 \mod17$ is $15$, so $k\equiv14*15=210\equiv6 \mod17$. So$ k=6 +17t$. So minimal $k=6$. Then $m=8 +17*6=8 +102=110$. Then why does $110^4 +1\equiv51 +136*6=51 +816 \mod289$. $816$ divided by $289$: $289*2=578$, $816-578=238$. So $51 +238=289\equiv0 \mod289$. Ah! So $51 +816=867$. $867$ divided by $289=3*289=867$. So $867\equiv0 \mod 289$. So $110^4 +1\equiv0 \mod 289$. Therefore, m=110 is indeed the solution. My initial confusion was due to miscalculating $110^4 +1$ directly, but since $110\equiv8 \mod17$, but in $\mod289$, the higher power terms do matter. So the calculation via Hensel's lemma was correct. Therefore, the minimal m is $110$.

But let me check if there's a smaller $m$. Suppose we take $k=6 -17= -11$. Then $m=8 +17*(-11)=8 -187= -179\equiv110 \mod289$. So negative solution corresponds to positive one. So the minimal positive solution is $110$.

Therefore, the answer is $p=17$, $m=110$. Wait, but the problem says "Find the least positive integer $m$ such that $m^4 +1$ is divisible by $p^2$." Since $p=17$, then $m=110$.

But before finalizing, let me check $m=110$: compute $110^4 +1$ divided by $17^2=289$. $110^4=146410000$. $146410000 +1=146410001$. Divide by $289$: Let's see $289*500000=144500000$. $146410001 -144500000=1910001$. $289*6000=1734000$. $1910001-1734000=176001$. $289*600=173400$. $176001-173400=2601$. $289*9=2601$. So total is $500000 +6000 +600 +9=506609$. So $289*506609=146410001$. Yes, so $110^4 +1=289*506609$, so divisible by $289$. Therefore, $m=110$ is correct.

But wait, maybe there's a smaller $m$? Let me check $m=110-289= -179$, which is not positive. The next lower positive would be $m=110-289*0=110$, which is the minimal positive solution. Therefore, the minimal positive integer $m$ is $110$.

Therefore, the answer is $\boxed{110}$. 
\end{trace}

\begin{answer}
To find the least prime number \( p \) such that there exists a positive integer \( n \) where \( n^4 + 1 \) is divisible by \( p^2 \), we start by considering primes \( p \) where \( -1 \) is a fourth power residue modulo \( p \). These primes are congruent to \( 1 \mod 8 \). The smallest such prime is 17.

Next, we need to verify if there exists an integer \( n \) such that \( n^4 + 1 \equiv 0 \mod 17^2 \). We know that \( 8^4 \equiv -1 \mod 17 \). Using Hensel's lemma, we lift this solution modulo \( 17^2 \).

Let \( m = 8 + 17k \). Expanding \( (8 + 17k)^4 \) modulo \( 289 \) (which is \( 17^2 \)) using the binomial theorem, we find the linear term coefficient and solve for \( k \):

\[
(8 + 17k)^4 + 1 \equiv 51 + 136k \mod 289
\]

We need \( 51 + 136k \equiv 0 \mod 289 \). Solving \( 136k \equiv -51 \mod 289 \), we find \( k \equiv 6 \mod 17 \). Thus, the smallest positive \( k \) is 6, leading to \( m = 8 + 17 \cdot 6 = 110 \).

Verification shows \( 110^4 + 1 \) is indeed divisible by \( 289 \). Therefore, the least positive integer \( m \) such that \( m^4 + 1 \) is divisible by \( 17^2 \) is \(\boxed{110}\).
\end{answer}

\end{document}